\documentclass{article}

\usepackage{microtype}
\usepackage{graphicx}
\usepackage{subfigure}
\usepackage{booktabs} 
\usepackage{caption}

\usepackage{hyperref}

\usepackage{xspace}
\newcommand{\alg}{MetaDiffuser\xspace}
\newcommand{\pdt}{Prompt-DT\xspace}

\def\gD{{\mathcal{D}}}

\def\gM{{\mathcal{M}}}
\def\gN{{\mathcal{N}}}

\def\gT{{\mathcal{T}}}

\newcommand{\tabincell}[2]{\begin{tabular}{@{}#1@{}}#2\end{tabular}}
\usepackage{multirow}
\usepackage{booktabs}
\usepackage{bbding}
\usepackage{xcolor}
\definecolor{mycolor}{rgb}{0.122, 0.435, 0.698}
\usepackage[most]{tcolorbox}
\usepackage{algorithm}
\usepackage{algorithmic}

\renewcommand{\algorithmicrequire}{\textbf{Input:}}
\renewcommand{\algorithmicensure}{\textbf{Output:}}

\usepackage{siunitx}

\usepackage[accepted]{icml2023}

\usepackage{amsmath}
\usepackage{amssymb}
\usepackage{mathtools}
\usepackage{amsthm}

\usepackage[capitalize,noabbrev]{cleveref}

\theoremstyle{plain}

\theoremstyle{definition}

\theoremstyle{remark}

\usepackage[textsize=tiny]{todonotes}

\usepackage{caption}
\usepackage{enumitem}

\icmltitlerunning{MetaDiffuser: Diffusion Model as Conditional Planner for Offline Meta-RL}

\begin{document}

\twocolumn[
\icmltitle{MetaDiffuser: Diffusion Model as Conditional Planner for Offline Meta-RL}



\icmlsetsymbol{equal}{*}

\begin{icmlauthorlist}
\icmlauthor{Fei Ni}{tju}
\icmlauthor{Jianye Hao}{tju,huawei}
\icmlauthor{Yao Mu}{hku}
\icmlauthor{Yifu Yuan}{tju}
\icmlauthor{Yan Zheng}{tju}
\icmlauthor{Bin Wang}{huawei}
\icmlauthor{Zhixuan Liang}{hku}
\end{icmlauthorlist}

\icmlaffiliation{hku}{Department of Computer Science, The University of Hong Kong, Hong Kong SAR}
\icmlaffiliation{tju}{College of Intelligence and Computing, Tianjin University, Tianjin, China}
\icmlaffiliation{huawei}{Huawei Noah’s Ark Lab, Beijing, China}

\icmlcorrespondingauthor{Jianye Hao}{jianye.hao@tju.edu.cn}

\icmlkeywords{Machine Learning, ICML}

\vskip 0.3in
]



\printAffiliationsAndNotice{}  

\begin{abstract}

Recently, diffusion model shines as a promising backbone for the sequence modeling paradigm in offline reinforcement learning~(RL). However, these works mostly lack the generalization ability across tasks with reward or dynamics change. To tackle this challenge, in this paper we propose a task-oriented conditioned diffusion planner for offline meta-RL~(MetaDiffuser), which considers the generalization problem as conditional trajectory generation task with contextual representation. The key is to learn a context conditioned diffusion model which can generate task-oriented trajectories for planning across diverse tasks. To enhance the dynamics consistency of the generated trajectories while encouraging trajectories to achieve high returns, we further design a dual-guided module in the sampling process of the diffusion model.
The proposed framework enjoys the robustness to the quality of collected warm-start data from the testing task and the flexibility to incorporate with different task representation method. The experiment results on MuJoCo benchmarks show that MetaDiffuser outperforms other strong offline meta-RL baselines, demonstrating the outstanding conditional generation ability of diffusion architecture. More visualization results are released on \href{https://metadiffuser.github.io}{project page}.

\end{abstract}

\section{Introduction}
\label{intro}

Offline Reinforcement Learning (Offline RL)~\citep{levine2020offline}  aims to learn policies from pre-collected data without interacting with the environment and has made many success in the fields of games~\citep{chen2021decision, li2022pmic}, robotic manipulation~\citep{ebert2018visual}, sequential advertising~\citep{hao2020dynamic}. However, one of the inherent difficulties of offline RL is the challenges to generalize to unseen tasks. Recent work in offline meta-RL ~\citep{mitchell2021offline,mbml,improved-context,focal,mu2022domino} aims to solve this problem by training a meta-policy from multi-task offline datasets that can efficiently adapt to unseen tasks with small amounts of warm-start data.

\begin{figure}[!t]
\centering
\includegraphics[scale=0.3]{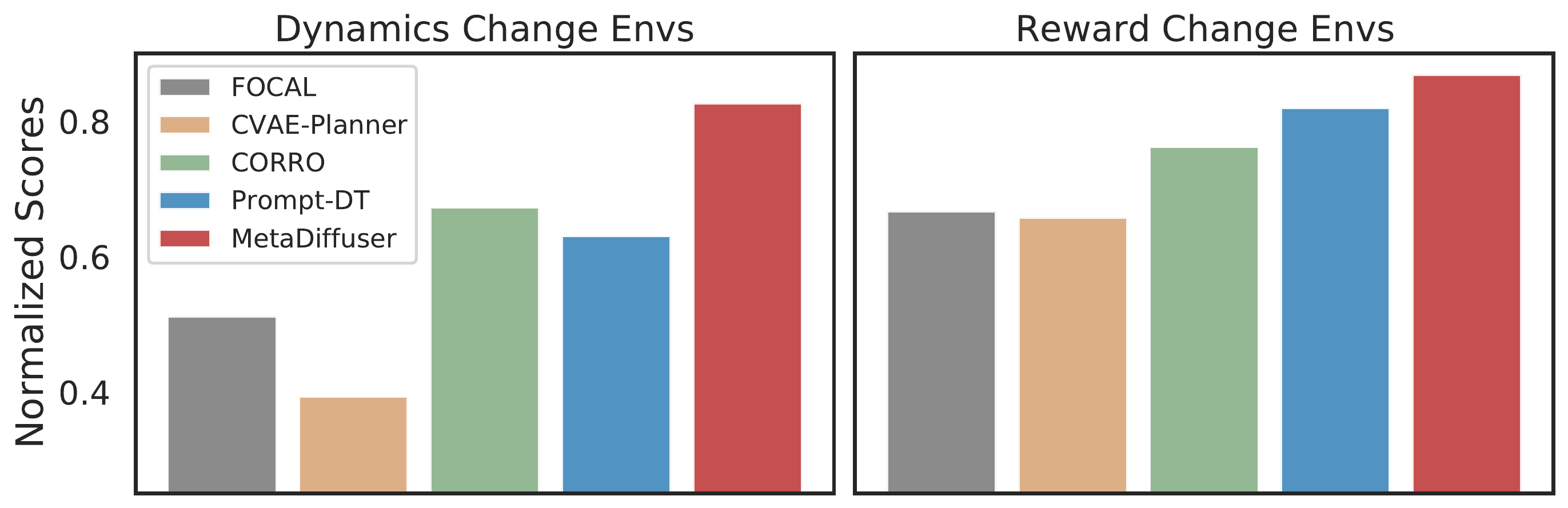}
\vspace{-20pt}
\caption{Overall few-shot generalization performance comparisons on various environments including 2 domains with dynamics change and 4 domains with reward change. The expert performance in each environment is chosen as normalized baseline. }
\vspace{-16pt}
\label{fig:performance_overall}
\end{figure}

Conventional offline meta-RL methods ~\cite{improved-context,yuan2022robust} learn a context encoder to infer task representation and a meta policy conditioned on the learned context for generalization across tasks. These works extended from the online meta-RL setting, still rely on context-conditioned policy trained by temporal difference (TD) learning, which may potentially cause instability in policy optimization and limited performance~\citep{levine2020offline, ajay2022conditional}. A more recent work \pdt~\citep{xu2022prompting} turn to tackle the generalization problem from the sequence modeling perspective, which joint models state-action trajectories to avoid TD-learning. This approach uses prompting method to generalize across unseen tasks without the need for explicit extraction of task representation through pre-trained context encoder. However, the key limitation is that the quality of the pre-collected warm-start data must be high enough, which is challenging to collect in unseen tasks,  to act as an expert prompt for guiding sequence generation, otherwise performance may suffer with random or medium data.  The aforementioned limitations raise a key question: Can we design a offline meta-RL framework to achieve the generalization ability across multiple tasks with robustness for the quality of warm-start data while utilize the promising ability of sequence-modeling paradigm?

Planning with diffusion model~\citep{janner2022planning} provides a promising paradigm for offline RL, which utilizes diffusion model as a trajectory generator by joint diffusing the states and actions from the noise to formulate the sequence decision-making problem as standard generative modeling. The concurrent works~\citep{ajay2022conditional, wang2022diffusion} also showcase the potential of the diffusion model as a highly promising generative model, highlighting its ability to serve as a key backbone for addressing sequence modeling problems in RL, while avoiding the limitations of TD-learning. But these works focus on a single task and lack research on generalization ability across tasks, which leaves the conditioned diffusion unexplored for offline meta-RL. 
However, conditioned diffusion model has made significant progress in vision and language tasks~\citep{ho2022classifier}, such as DALL-E~\citep{ramesh2022hierarchical} and ImageGen~\citep{saharia2022photorealistic} for text-to-image generation tasks. These works demonstrate the powerful conditional generation capabilities of conditioned diffusion models with the textual label without the need for expert images as prompts.

Inspired by this, we propose a novel framework for offline meta-RL, named \alg that leverages diffusion model to conduct desired trajectories generation for generalization across unseen tasks. 
During meta-training, to provide accurate conditional labels for subsequent trajectory generation, we first pre-train an accurate context encoder that can capture 
task-relevant information
from offline trajectories mixed with different tasks. 
Then the compact task representation 
is injected as a contextual label to the conditional diffusion model to manipulate the task-oriented trajectories generation. In this way, the diffusion model learns to estimate the conditional distribution of multi-task distributions based on the  
task-oriented 
context. During meta-testing, with the predicted context from provided warm-start data in the testing task, the conditional diffusion model can denoise out desired trajectories for the testing task. The generated trajectories can guide the subsequent action to step into the next state, similar to the planning~\citep{yuan2023euclid} in RL. Moreover, to decrease the discrepancy between generated trajectories and real-rollout trajectories, we design an effective dual-guide to enhance the dynamics consistency of generated trajectories while encouraging the high return simultaneously. 
The contributions of this work are as follows:
\begin{itemize}[leftmargin=*,itemsep=2pt,topsep=0pt,parsep=0pt]
\item \textbf{Generalization Ability}: We
propose \alg to leverage the diffusion model to conduct conditional trajectory generation to achieve the generalization ability across unseen tasks.
\item \textbf{Robustness and Flexibility}: \alg enjoys the flexibility to incorporate with different task representation method and the robustness to the quality of collected warm-start data at the testing task.
\item \textbf{Dual-guide Enhanced Planner}: 
We design the dual-guide of both dynamics and rewards to ensure the feasibility of guided trajectories while encouraging the generated trajectories to achieve high returns. 
\item \textbf{Superior Performance}: The experiments on various benchmarks empirically show that \alg much better generalizes to unseen tasks than prior methods. 
\end{itemize}

\section{Related Work}
\subsection{Offline Meta-RL}
Offline meta-RL investigates learning to learn from offline data, with the aim to quickly adapt to unseen tasks. Recent works~\citep{mitchell2021offline,mbml,improved-context}, including FOCAL~\citep{focal} and CORRO~\citep{yuan2022robust}, trains a context encoder for compact task representation for the conditioned policy to generalize.  These methods extended from the traditional online meta-RL setting, still rely on context-conditioned policy trained by TD-learning, which may potentially cause instability in policy optimization and limited performance. 
\pdt~\citep{xu2022prompting} turns to solve the generalization problem from the sequence modeling perspective, which joint models state-action trajectories to avoid TD-learning. This approach can utilize the collected prompt as a prefix to generalize across tasks without the need for explicit context encoder. However, the key limitation is the high requirement for the quality of warm-start data as prompt, which is challenging to pre-collect in unseen task. See more discussion in \cref{app:additional discussion}.
To combine the best of both context-based manner and sequence-modeling fashion, we propose \alg, which not only avoiding TD-loss, but also enjoy the robustness to the quality of warm-start data.

\subsection{Diffusion Model for Sequence Decision Making}

Recently, many works have emerged to utilize diffusion models to solve sequence decision-making tasks, showing the great potential of diffusion model as a promising backbone of sequence modeling. Diffuser~\citep{janner2022planning} applies a diffusion model as a trajectory generator, which is trained by diffusing over the full trajectory of state-action pairs from the noises.
A separate reward model is trained to predict the cumulative rewards of each trajectory sample, then the gradient guidance from reward model is injected into the reverse sampling stage. Then the first action in the generated trajectories will be applied to execute in the environment to step into the next state, which repeats in a loop until the terminal.
The consequent work Decision Diffuser~\citep{ajay2022conditional} frames offline sequential decision-making as conditional generative modeling based on returns, constraints and skills to eliminate the complexities in traditional offline RL.
The concurrent work Diffusion-QL~\citep{wang2022diffusion}, build policy with the reverse chain of a conditional diffusion model,
which allows for a highly expressive policy class,
 as a strong policy-regularization method. 
However, these works mostly focus on a single task and lack the generalization ability to unseen tasks in the setting of offline meta-RL. Our approach \alg leverages the conditioned diffusion model to conduct conditional trajectory generation to achieve the generalization across unseen tasks with different reward functions or dynamics.

\subsection{Conditional Diffusion Model}

Recently, there have been incredible advances in the field of conditional content generation with the strong generation capabilities of conditioned diffusion models.
Conditional diffusion model pushes the state-of-the-art on 
text-to-image generation tasks such as DALL-E~\citep{ramesh2022hierarchical} and ImageGen~\citep{saharia2022photorealistic}.
The technique of conditioning can divide into two fashions: classifier-guided~\citep{nichol2021improved} and classifier-free~\citep{ho2022classifier}. The former improves sample quality while reducing diversity in conditional diffusion models using gradients from a pre-trained classifier $p_{\phi}(\boldsymbol{y}|\boldsymbol{x}_{k})$ during sampling. The latter is an alternate technique that avoids this pre-trained classifier by instead jointly training a single diffusion model on conditional $\epsilon_{\theta}(\boldsymbol{x}_{k}, \boldsymbol{y}, k)$ and unconditional $\epsilon_{\theta}(\boldsymbol{x}_{k}, k)$ noise model via randomly dropping conditional label $\boldsymbol{y}$. 

In fact, the aforementioned Diffuser~\citep{janner2022planning} can also be considered as a classifier-guided conditional diffusion model, where the pre-trained reward model is another form of classifier for evaluating the sample quality. Our designed \alg builds upon the Diffuser and additionally incorporates classifier-free manner, by injecting the context as label $\boldsymbol{y}$ into the conditional noise model $\epsilon_{\theta}(\boldsymbol{x}_{k}, \boldsymbol{y}, k)$, achieving more precise conditional generation. The details about the relationship between two different conditional fashions can be found in \cref{appedix:classifier}.

\section{Preliminaries}
\subsection{Problem Formulation}

The reinforcement learning problem can be generally modeled as a Markov Decision Process (MDP), represented as $\gM=(\mathcal{S}, \mathcal{A}, T, \rho, R)$, where $\mathcal{S}$ is the state space, $\mathcal{A}$ is the action space, $T(s'|s,a)$ is the transition dynamics of the environment, $\rho(s)$ is the initial state distribution, $R(s,a)$ is the reward function.
The objective is to find a policy $\pi(a|s)$ that optimizes the expected cumulative rewards, $\mathbb{E}_{s_0 \sim \rho, \pi} \sum_t\gamma^t R(s_t)$, starting from the initial state.
In the offline meta-RL setting, aiming to adapt to new tasks via pre-collected data quickly, an agent is given a set of tasks $\gT$, where a task $\gT_i \in \gT$ is defined as $(\gM_i, \pi_i)$, containing an MDP $\gM_i$ and a behavior policy $\pi_i$.
For each task $\gT_i$, the agent is provided with a pre-collected dataset $\gD_i$, which contains trajectories sampled using $\pi_i$.
The agent is trained with a subset of training tasks denoted as $\gT^{train}$ and is expected to find the optimal policies in a set of test tasks $\gT^{test}$, which is disjoint with $\gT^{train}$.

\subsection{Diffusion Model}
Diffusion models \citep{sohldickstein2015nonequilibrium, ho2020denoising} are a type of generative model that
consists a forward diffusion process and a reverse denoising process to learn the data distribution $q(\boldsymbol{x})$. 
Here, the data-generating procedure is modelled with a predefined forward noising process $q(\boldsymbol{x}_{k+1}|\boldsymbol{x}_{k}) \coloneqq \mathcal{N}(\boldsymbol{x}_{k+1}; \sqrt{\alpha_{k}}\boldsymbol{x}_{k}, (1-\alpha_{k})\boldsymbol{I})$ and a trainable reverse process $p_{\theta}(\boldsymbol{x}_{k-1}|\boldsymbol{x}_{k}) \coloneqq \mathcal{N}(\boldsymbol{x}_{k-1}|\mu_{\theta}(\boldsymbol{x}_{k}, k), \Sigma_{k})$, where $\gN(\mu,\Sigma)$ denotes a Gaussian distribution with mean $\mu$ and variance $\Sigma$, $\alpha_k \in \mathbb{R}$ determines the variance schedule, $\boldsymbol{x}_{0} \coloneqq \boldsymbol{x}$ is a sample, $\boldsymbol{x}_{k}$ are the sequentially sampled latent variables for $k=1,\dots,K$, and $\boldsymbol{x}_{K} \sim \mathcal{N}(\boldsymbol{0}, \boldsymbol{I})$ for carefully chosen $\alpha_{k}$ and long enough $K$. Starting with Gaussian noise, samples are then iteratively generated through a series of reverse denoising steps by the predicted noise. The predicted noise $\epsilon_\theta(\boldsymbol{x}_{k}, k)$, parameterized with a deep neural network, estimates the noise $\epsilon \sim \gN(0,I)$ added to the dataset sample $\boldsymbol{x}_{0}$ to produce noisy $\boldsymbol{x}_{k}$, which can be trained by a simplified surrogate loss~\citep{ho2020denoising}:
$ \mathcal{L}_{\text{denoise}}(\theta) \coloneqq \mathbb{E}_{k \sim [1,K], \boldsymbol{x}_{0} \sim q, \epsilon \sim \mathcal{N}(\boldsymbol{0}, \boldsymbol{I})}[||\epsilon - \epsilon_{\theta}(\boldsymbol{x}_{k}, k)||^{2}]$.

\section{Methodology}

\begin{figure*}[t]
    \centering

    \includegraphics[width=0.9\textwidth]{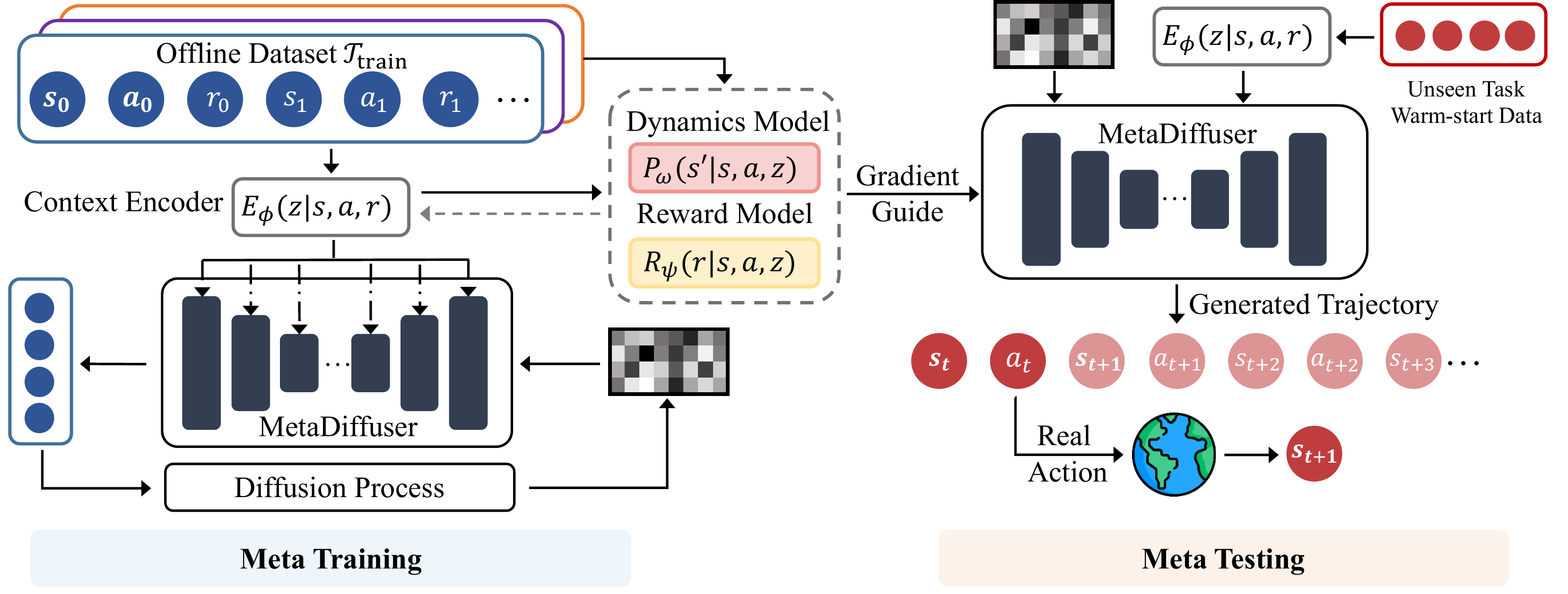}

    \caption{The overview of \alg. During \textbf{meta-training} phase, a task-oriented context encoder is trained jointly with conditioned dynamics model and reward model in a self-supervised manner to infer the current task from the recent historical transitions. Then, the multi-task trajectories can be labeled with the trained context encoder and the inferred context are injected in the conditioned diffusion model to estimating the multi-modal distribution mixed by different training tasks. During \textbf{meta-testing} phase, context encoder can capture the task information from provided warm-start data from the test task. Then the conditioned diffusion model can manipulate the noise model to denoise out desired trajectories for the test task with the inferred context. Additionally, the pretrained dynamics model and reward model can serve as classifiers for evaluation, with gradient to guide the conditional generation in a classifier-guide fashion. 
    }
    \label{fig:framework}
    \vspace{-12pt}
\end{figure*}

To tackle the generalization challenge from the sequence modeling perspective,
we propose \alg, a novel offline meta-RL framework, which leverages the conditioned diffusion model to conduct task-oriented trajectories generation for generalization across unseen tasks.
As shown in ~\cref{fig:framework}, the overall generalization process can explicitly be divided into meta-training and meta-testing. During \textbf{meta-training}, in order to provide accurate conditional labels for subsequent trajectory generation, we first need to pre-train an accurate context encoder that can capture both reward changes and dynamics changes from trajectories. Then the compact task representation inferred by the context encoder is injected as a contextual label into the step-wise denoising process from the Gaussian noise for estimating the conditional distribution of multi-task trajectories. During \textbf{meta-testing}, with predicted context from provided warm-start data, the conditional diffusion model can denoise out desired trajectories for the testing task. Moreover, to alleviate the discrepancy between generated trajectories and real-rollout trajectories, the previously trained reward model and dynamics model are utilized as a trajectory evaluator to enhance the dynamics consistency and high returns of trajectories.

\subsection{Task-oriented Context Encoder}
\label{sec:context encoder}
To manipulate the conditional trajectory generation with a high correlation with the desired specific task, it is necessary to establish an accurate mapping from trajectories to the contextual label it belongs to.
Considering the environments in the meta-RL setting can change in reward functions and transition dynamics, we expect the context can fully distinguish between the two types of environmental changes with a unified learning objective. 
For this, we propose a simple yet effective context encoder $E_{\phi}$, jointly with generalized reward model $R_\psi$ and dynamics model $P_\omega$. We augment context into state and action to minimize the prediction loss of both dynamics and reward simultaneously.

Specifically, given the multi-task offline dataset $\gD$, which contains the trajectories $\tau^\gM =\{(s_t,a_t,r_t,s_{t+1})\}_{t=1}^K$ with horizon $K$ for training task $\gM \sim \gT^{train}$. 
For each trajectories, a trajectories segment $\tau^\gM_t =\{(s_{t+i},a_{t+i},r_{t+i},s_{{t+i+1}})\}_{i=0}^h$  of size $h$ are sampled started from random selected $t$. With the historical sub-trajectories, the context encoder $E_{\phi}$ captures the latent representations $z_t = E_{\phi}(\tau^\gM_t)$ as the contextual information of the task. Then the generalized reward model $R_\psi$ and dynamics model $P_w$ are conditioned on $z$, parameterized with $\psi, \omega$. The context encoder is trained jointly by  
minimizing the state transition and reward prediction error  conditioned on the learned context:
\begin{equation}
\footnotesize
\label{eq:pretrain context encoder}
\vspace{-10pt}
\begin{aligned}
\mathcal{L}_{\phi,\psi,\omega}&= -\mathbb{E}_{\left(s_t, a_t, r_t, s_{t+1}\right) \sim \tau^\gM_t;\gM \sim \gT^{train} } \Big[ \mathbb{E}_{z_t=E_\phi(z_t \mid \tau^\gM_t)} \\
& \left[\log P_\omega(\hat{s}_{t+1}|s_t, a_t, z_t)+ \log R_\psi(\hat{r}_t|s_t, a_t, z_t) \right] \Big]
\end{aligned}
\end{equation}

Moreover, our method additionally trains the generalized reward model and dynamics model as byproducts, which will play a key role as a useful classifier in the later classifier-guided conditional generation module. It should be noted that our framework is flexible to other representation methods, the further analysis is illustrated in \cref{sec:representation}. The detailed experimental results about distribution shift of the quality of training data can be found in \cref{abla: context data quality}.

\subsection{Conditional Diffusion Architecture}
\label{sec:conditional diffusion architecture}
Inspired by the great success of the diffusion model in text-to-image tasks, which generates images based on text labels from noises, we leverage the diffusion model as a trajectory generator conditioned on the task-oriented context. 
Following Diffuser~\citep{janner2022planning}, the states and actions in the trajectory are generated simultaneously per time step $t$ over the planning horizon $H$:
\begin{equation}
\footnotesize
    \boldsymbol{x}_{k}(\tau) = (s_{t},a_t, s_{t+1},a_{t+1} ..., s_{t+H-1},a_{t+H-1})_{k} 
\end{equation}
where $k$ denotes the timestep in the denoising process.
Now we have the pre-trained context encoder to infer the task labels for different tasks, we can additionally condition the diffusion process on the contextual information of the tasks. 
In this way, we formulate the meta-RL problem as the conditional generative modeling problem:

\begin{equation}
\footnotesize
\label{eq:cond_gen_model}
\theta^*= \arg \max_\theta \mathbb{E}_{\tau \sim \gD}[\log p_\theta(\boldsymbol{x}_{0}(\tau)|\boldsymbol{y} = E_\phi(\tau))]
\end{equation}

where the conditional label $y$ denotes the task-oriented context inferred from the pre-collected offline data from the current task by context encoder $E_{\phi}$. The goal is to estimate the conditional data distribution with $p_{\theta}$ so we can later generate desired trajectory $\boldsymbol{x}_0(\tau)$ according to the context label from unseen tasks.
The forward diffusion process $q$ and the reverse denoising process $p_\theta$ can be formulated as:
\begin{equation}
\footnotesize
\label{eq:diff_plan}
    q(\boldsymbol{x}_{k+1}(\tau)|\boldsymbol{x}_{k}(\tau)), \;\;\;\; p_{\theta}(\boldsymbol{x}_{k-1}(\tau)|\boldsymbol{x}_{k}(\tau), \boldsymbol{y} = E_\phi(\tau)))
\vspace{-2pt}
\end{equation}
Specifically, for each trajectory $\tau$ in the training 
 offline dataset, we first sample a Gaussian noise $\epsilon \sim \mathcal{N}(\boldsymbol{0}, \boldsymbol{I})$ and a denoising timestep $k \in \{1,\ldots,K\}$. Then we construct a noisy array with the same dimension of $\boldsymbol{x}_{k}(\tau)$ and finally predict the denoising noise as $\hat{\epsilon}_{\theta} =  \epsilon_{\theta}(\boldsymbol{x}_{k}(\tau), \boldsymbol{y}(\tau), k)$ 
 in the denoising step $k$. 
 
For the classifier-free conditioned diffusion model~\citep{ho2022classifier}, the commonly used technique is to randomly drop out the conditioning for improving the quality of generated samples. Intuitively, we also train the noise model jointly with a single diffusion model on conditional and unconditional objective via randomly dropping the conditioning context label with probability $\beta$. The proper drop probability can balance off the diversity and the relevance of the context label of generated trajectories. The detailed analysis about the effects of different context drop probability can be found in \cref{abla:drop}.

So far, with the mixed trajectories datasets $\mathcal{D}$ paired with contextual information of the task it belongs to, we can train the reverse denoising process $p_{\theta}$, parameterized through the conditional noise model $\epsilon_\theta$ with the following loss:
\begin{equation}
\footnotesize
\vspace{-4pt}
\label{eq:conditioned diffusion loss}
\begin{aligned}
    \mathcal{L}(\theta)=\mathbb{E}_{k, \tau \in \mathcal{D}} \left[\left\|\epsilon-\epsilon_\theta\left(\boldsymbol{x}_k\left(\tau\right),(1-\beta) E_\phi\left(\tau\right)+\beta\varnothing, k\right)\right\|^2\right]
\end{aligned}
\end{equation}

After training a conditioned diffusion model for imitating expert trajectories in the offline datasets, we now discuss how to utilize the trained diffusion model to achieve the generalization across unseen tasks.
During meta-testing, the context encoder captures the task information context from pre-collected trajectories as warm-start data and infer the
task-oriented context as the conditional label $\boldsymbol{y} = E_\phi(\tau)$. Then the context label can be injected into conditioned diffusion model to guide the desired expert trajectory generation for the current task. $\boldsymbol{x}_{0}(\tau)$ is sampled by starting with Gaussian noise $\boldsymbol{x}_{K}(\tau)$ and refining $\boldsymbol{x}_k(\tau)$ into $\boldsymbol{x}_{k-1}(\tau)$ at each intermediate timestep with the perturbed noise:
\begin{equation}
\vspace{-2pt}
\label{eq:denosing process 1}
    \hat{\epsilon} = \omega \epsilon_\theta(\boldsymbol{x}_k(\tau), \boldsymbol{y}, k) + (1 - \omega)\epsilon_\theta(\boldsymbol{x}_k(\tau), \varnothing, k)
\end{equation}

where the scalar $\omega$ denotes
the guidance weight in the classifier-free conditioned diffusion model. Setting $\omega = 1$ disables classifier-free guidance while
increasing $\omega > 1$ strengthens the effect of guidance.
Based on the 
context-conditioned
noise generated iteratively, 
the desired trajectories containing future states and actions can be denoised from the noise step-wisely.
With the generated trajectories, the first action will be applied to execute in the environment to step into the next state. This procedure repeats in a standard receding-horizon control loop, similar to traditional planning in RL, described in \cref{app:pesudocodes}.  For architecture details, please refer to \cref{app:hyperparam}.

\subsection{Dual-guide Enhanced Planner}
\label{sec:dual guide}

\begin{figure}[t]
    \begin{center}
    \centerline{\includegraphics[width=\linewidth]{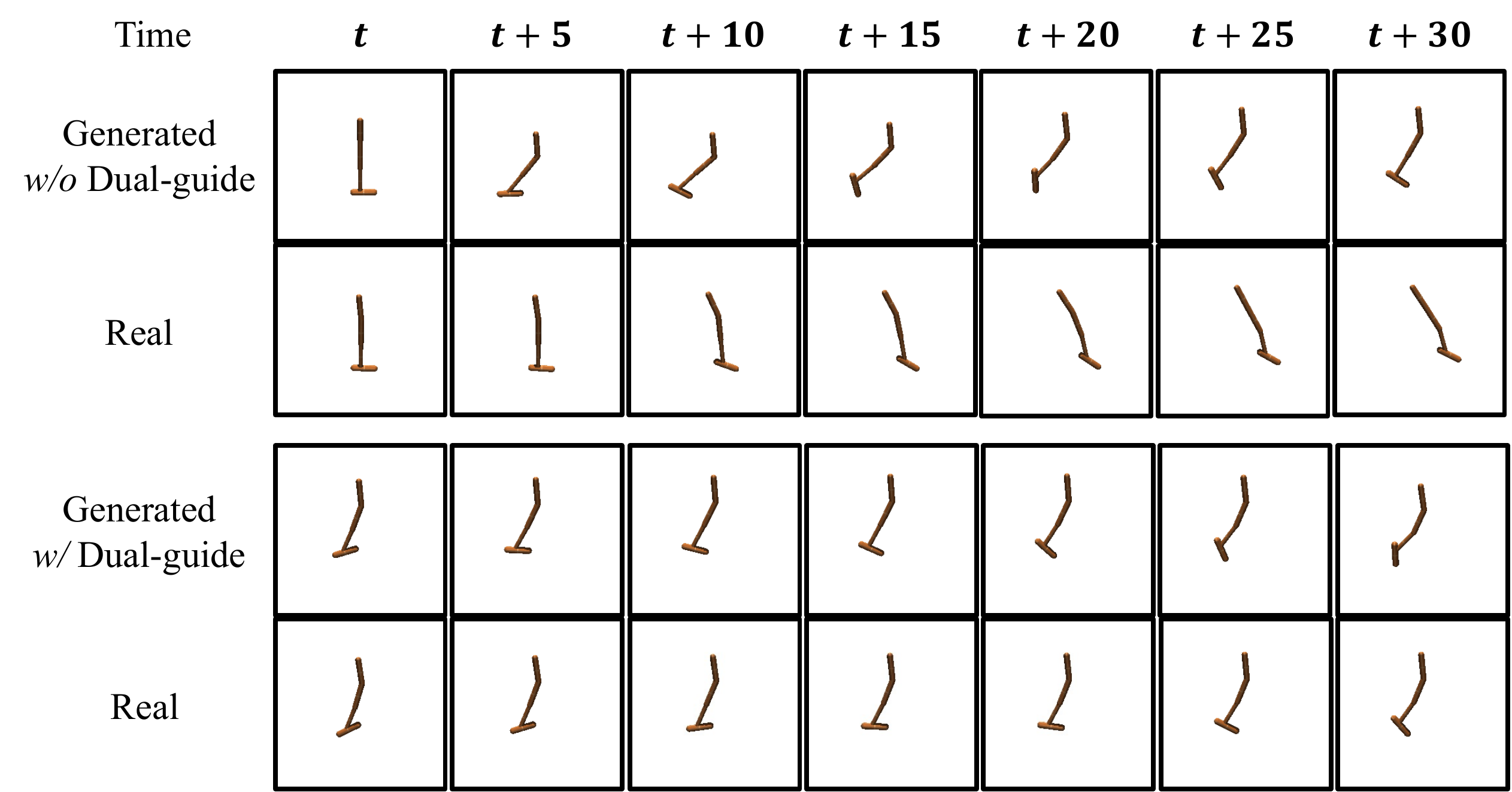}}
    \vspace{-8pt}
    \caption{The visualization of an extreme case about generated trajectories and real trajectories rollout according to actions within generated trajectories in Hopper-Param, as an environment with dramatic dynamics changes. With dual-guide, the generated trajectories are less aggressive in expected rewards and more dynamics consistent to enhance the reachability between adjacent states.
    }
    \label{fig:pose_vis}
    \end{center}
    \vspace{-20pt}
\end{figure}

Previous work~\citep{janner2022diffuser} trains an extra reward predictor $\mathcal{J}$ to evaluate the accumulative return of generated trajectories and utilizes the gradient of return as a guidance in the sampling process of diffusion model,  to encourage the generated trajectories to achieve high return.
However, during meta-testing for unseen tasks, as shown in the top part of \cref{fig:pose_vis},  the conditional generated trajectories may not always obey dynamics constraints due to the aggressive guidance aim for high return,  making it difficult for the planner to follow the expected trajectories during the interaction with the environment. Therefore, we propose a dual-guide  to
enhance the dynamics consistency of generated trajectories while encouraging the high return $\mathcal{J}$ simultaneously. 

For this,  we utilize the previously pretrained dynamics model,  to predict the future state of the generated trajectory based on its planned actions, then compared it to the states in the generated trajectory. The dynamics discrepancy $\zeta$ serves as an important metric to evaluate the consistency and reachability of the generated trajectory. Then the gradient from dual-guide can be formulated as:
\begin{equation}
\footnotesize
\begin{split}
 &g = \nabla \mathcal{J}(\boldsymbol{x}_{k}(\tau)) + \lambda\nabla \mathcal{\zeta}(\boldsymbol{x}_{k}(\tau)) \\
&\mathcal{J}\left(\boldsymbol{x}_k(\tau)\right)=\sum_{t=0}^T R_\psi\left(s_t, a_t, z_t\right)\\
&\zeta\left(\boldsymbol{x}_k(\tau)\right)=\sum_{t=0}^T\left\|s_{t+1}-P_\omega\left(\hat{s}_{t+1} \mid s_t, a_t, z_t\right)\right\|^2\\
\end{split}
\vspace{-8pt}
\label{eq:gradient}
\end{equation}

where $\lambda$ denotes the relative scaling coefficient between the dynamics guide and reward guide to balance off the high reward and low discrepancy. The detailed ablation study about the scaling effect can be found in \cref{abla:dual}. The visualization of an intuitive example is shown in \cref{fig:pose_vis}.

In this way, incorporate \alg not only conducts the classifier-free manner by injecting the context as label $\boldsymbol{y}$ into the conditional noise model $\epsilon_{\theta}(\boldsymbol{x}_{k}, \boldsymbol{y}, k)$, achieving more precise conditional generation, but also incorporate the classifier-guide fashion in Diffuser where the single reward guide is expanded to designed dual-guide for more complex environment change in meta-RL setting. 
Formally, the denoising process in \cref{eq:denosing process 1} can be extended as:
\begin{equation}
\footnotesize
\begin{aligned}
\label{eq:denosing process 2}
    \hat{\epsilon} & \coloneqq  \underbrace{\omega \epsilon_\theta(\boldsymbol{x}_k(\tau), E_\phi(\tau), k) + (1 - \omega)\epsilon_\theta(\boldsymbol{x}_k(\tau), \varnothing, k)}_{\text{classifier-free}} \\
    & - \underbrace{\sqrt{1-\bar{\alpha}_t}  \nabla _{\boldsymbol{x}_{k}(\tau)} \Big[\mathcal{J}(\boldsymbol{x}_{k}(\tau)) + \lambda \mathcal{\zeta}(\boldsymbol{x}_{k}(\tau))\Big]}_{\text{classifier-guided}}
\end{aligned}
\end{equation}
The details about the relationship between two different conditional fashions can be found in \cref{appedix:classifier}.

\begin{table*}[t]
\caption{Average test returns of \alg against other baselines with a few-shot manner. }
\label{table:averge return}
\centering
\footnotesize
\begin{tabular}{ccccccc}
\toprule
\textbf{Environment}   & \textbf{FOCAL} & \textbf{CVAE-Planner} & \textbf{CORRO} &\textbf{Prompt-DT} &\textbf{\alg} &\textbf{Oracle} \\ \midrule

\textbf{\tabincell{c}{{Point-Robot}}}
& -5.99\scriptsize{$\pm 0.26$} & -6.11\scriptsize{$\pm 0.71$} & -5.59\scriptsize{$\pm 0.57$} & -5.04 \scriptsize{$\pm 0.35$} & \textbf{-4.48}\scriptsize{$\pm 0.28$} &  -3.97\scriptsize{$\pm 0.12$} \\

\textbf{\tabincell{c}{{Ant-Dir}}} 
 
& 151.3\scriptsize{$\pm 24.6$} & 130.6\scriptsize{$\pm 41.3$} & 193.3\scriptsize{$\pm 32.1$} & 213.2\scriptsize{$\pm 29.1$} & \textbf{247.7}\scriptsize{$\pm 16.8$} & 314.4\scriptsize{$\pm 9.2$} \\ 

\textbf{\tabincell{c}{{Cheetah-Dir}}} 

& {680.9}\scriptsize{$\pm 46.6$} & 759.4\scriptsize{$\pm 47.2$} & 823.5\scriptsize{$\pm 37.0$} & {931.7}\scriptsize{$\pm 21.3$} & \textbf{936.2}\scriptsize{$\pm 17.9$} & 943.4\scriptsize{$\pm 15.4$}\\ 

\textbf{\tabincell{c}{{Cheetah-Vel}}} 

& {-82.5}\scriptsize{$\pm 7.0$} & -87.1\scriptsize{$\pm 13.1$} & -56.2\scriptsize{$\pm 9.4$} & {-51.3}\scriptsize{$\pm 4.9$} & \textbf{-45.9}\scriptsize{$\pm 4.1$} & -32.4\scriptsize{$\pm 1.8$}\\ 
\midrule

\textbf{\tabincell{c}{{Walker-Param}}} 
 
& 245.6\scriptsize{$\pm 37.8$} & 187.9\scriptsize{$\pm 49.7$} & 300.5\scriptsize{$\pm 34.2$} & 287.7 \scriptsize{$\pm 32.1$} & \textbf{368.3}\scriptsize{$\pm 30.6$}  & 447.2 \scriptsize{$\pm 9.7$}\\ 

\textbf{\tabincell{c}{{Hopper-Param}}} 

& 203.6\scriptsize{$\pm 46.6$} & 157.3\scriptsize{$\pm 54.5$} & 289.3\scriptsize{$\pm 24.7 $} & 
265.2\scriptsize{$\pm 37.1$}& \textbf{356.4}\scriptsize{$\pm 16.9 $} &
429.6\scriptsize{$\pm 13.1 $}


\\ \bottomrule 

\end{tabular}
\end{table*}

\section{Experiments}
We conduct experiments on various tasks to evaluate the few-shot generalization performance of the proposed \alg. 
We aim to empirically answer the following questions:
1) Can \alg achieve performance gain on few-shot policy \textbf{generalization} compared to other strong baselines?
2) Can \alg show \textbf{robustness} to the quality of warm-start data?
3) Can \alg be a \textbf{flexible} framework to incorporate with any context representation method?

\subsection{Environments Settings}
We adopt a 2D navigation environment Point-Robot and multi-task MuJoCo control tasks to make comparisons, as classical benchmarks commonly used in meta-RL~\citep{mitchell2021offline,mbml,improved-context}. More details about   environments are available in \cref{app:benchmark}.
For each environment, different tasks are randomly sampled from the task distribution, divided into a training set $\gT^{train}$ and testing set $\gT^{test}$. On each task, we use SAC \cite{sac} to train a single-task policy independently. The trajectories of expert policy for each task are collected to be the offline datasets. See more details in \cref{app:implementation}.

\subsection{Baselines}
\textbf{FOCAL} \cite{focal} proposes a novel
negative-power distance metric learning method to train the context encoder for task inference, as an end-to-end offline meta-RL algorithm with high efficiency.

\textbf{CORRO} \cite{yuan2022robust} proposes a contrastive learning framework for task representations that are robust to the distribution mismatch of behavior policies in training and testing. CORRO demonstrates superior performance than prior context-conditioned policy-based methods.

\textbf{Prompt-DT} \cite{xu2022prompting} leverages the sequential modeling ability of the Transformer architecture and the prompt framework to achieve few-shot adaptation in offline RL, as a strong meta-RL baseline in sequence modeling fashion.

\textbf{CVAE-Planner} To investigate the influence of different generative architectures, we substitute the conditioned diffusion to conditioned VAE, serving the same role as a trajectory generator to guide the planning across tasks.

\subsection{The Generalization Ability on Task Adaptation}
To evaluate the performance on task adaptation, we sample tasks from the test set with warm-start data, which is pre-collected by a random policy or an expert policy.
Then we measure the few-shot generalization ability of different methods with the average episode accumulated reward. For fairness, all methods are trained with the same expert dataset in each environment to investigate whether the diffusion model facilitates few-shot generalization and the performance of \alg.

\begin{figure}[htb]
    \centering
    \centerline{\includegraphics[width=0.995\linewidth]{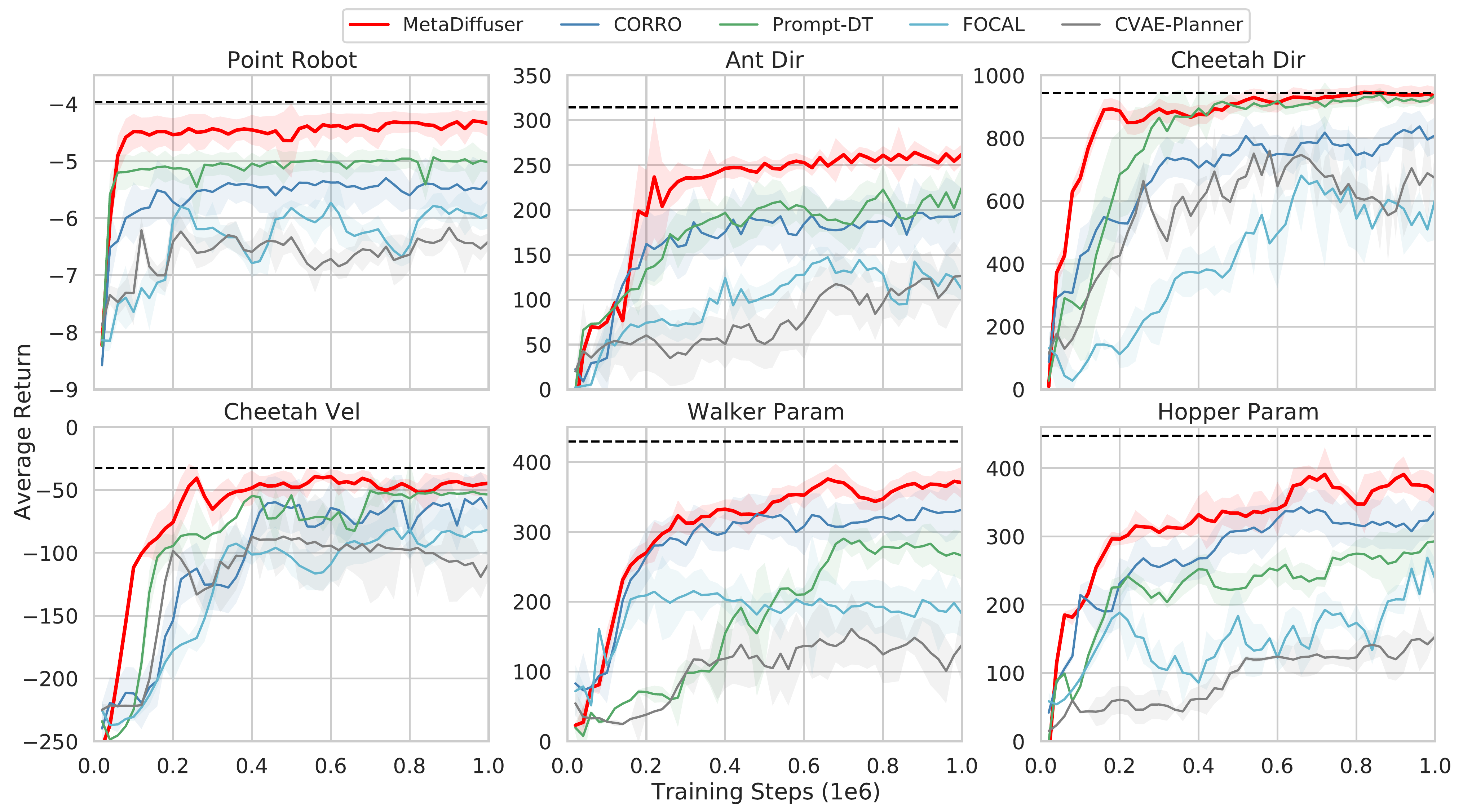}}
    \vspace{-10pt}
    \caption{\textbf{Meta-testing average performance} of \alg against baselines run over five random seeds in unseen tasks. 
    The dashed lines denote the oracle performance of expert policy trained separately for each test task. 
    }
    \label{fig:main_comparison_expert}
    \vspace{-16pt}
\end{figure}

The testing curves and converged performance are summarized in Figure~\ref{fig:main_comparison_expert} and Table~\ref{table:averge return} respectively, which contain six environments varying in dynamics and rewards.  
In relatively simple environments such as Point-Robot and Cheetah-Dir, \alg and \pdt significantly outperforms other baselines to a large extent. In Ant-Dir \alg outperforms other baselines by a large margin, which show the strong generalization ability in unseen task with different reward functions. Moreover, in Cheetah-Vel, \alg is more data-efficient to achieve better asymptotic performance than others, with the benefit of the strong generative capacity of the diffusion model. In dynamics change environments, such as Hopper-Param and Walker-Param, 
CORRO, as a context-based method, can have a more stable improvement than \pdt. The potential reason may be that the complex environment varying in dynamics is more challenging for \pdt to implicitly capture the dynamics information within a prompt.

\alg can outperform CORRO benefiting from the stability of the sequence-modeling framework instead of TD-learning.  The detailed analysis of the context representation method can be found in \cref{sec:representation}.The CVAE-Planner struggles to generalize to different tasks, illustrating the strong modeling capability of the diffusion model against CVAE, when meeting with the extreme multi-modal distribution.
We will illustrate the detailed analysis in \cref{sec:cvae}.

\begin{table*}[t]
\renewcommand\arraystretch{0.95}
\caption{The comparisons of the performance of Prompt-DT and \alg with different qualities of provided warm-start data during the meta-testing phase. The \textcolor{blue}{$\downarrow$ } denotes the performance drop with other quality of data.} 

\label{table:robustness}
\centering
\begin{small}
\begin{tabular}{@{}ccccccc@{}}
\toprule
\multicolumn{1}{c}{\multirow{2}{*}{\textbf{Environment}}} & \multicolumn{3}{c}{\textbf{\alg}} & \multicolumn{3}{c}{\textbf{Prompt-DT}}  \\ \cmidrule(lr){2-7} 
\multicolumn{1}{c}{}  & \textit{Expert} & \textit{Medium} &  \textit{Random} & \textit{Expert} & \textit{Medium} &  \textit{Random} \\ \midrule

\textbf{\tabincell{c}{{Point-Robot}}}
& \textbf{-4.48}\scriptsize{$\pm 0.28$} & -4.54\scriptsize{$\pm 0.31$} 
\textcolor{blue}{($\downarrow$ 1.5\%)}
& -4.61\scriptsize{$\pm 0.21$} \textcolor{blue}{($\downarrow$ 3.2\%)} & \textbf{-5.04}\scriptsize{$\pm 0.35$} 
& -5.17\scriptsize{$\pm 0.29$} \textcolor{blue}{($\downarrow$ 3.8\%)} & -5.85\scriptsize{$\pm 0.32$} \textcolor{blue}{($\downarrow$ 23.4\%)} 
  \\

\textbf{\tabincell{c}{{Ant-Dir}}} 
& \textbf{247.7}\scriptsize{$\pm 16.8$} & {238.9}\scriptsize{$\pm 18.1$} \textcolor{blue}{($\downarrow$ 3.6\%)}
& 213.8\scriptsize{$\pm 26.5$} \textcolor{blue}{($\downarrow$ 13.7\%)} & \textbf{213.2}\scriptsize{$\pm 29.1$} 
& {154.7}\scriptsize{$\pm 39.5$} \textcolor{blue}{($\downarrow$ 27.4\%)} & {40.1}\scriptsize{$\pm 16.3$} \textcolor{blue}{($\downarrow$ 81.2\%)} 
 \\ 

\textbf{\tabincell{c}{{Cheetah-Dir}}} 
& \textbf{936.2}\scriptsize{$\pm 17.9$} & 930.3\scriptsize{$\pm 18.5$} \textcolor{blue}{($\downarrow$ 0.6\%)} 
& 916.7\scriptsize{$\pm 21.8$} \textcolor{blue}{($\downarrow$ 1.9\%)} & \textbf{931.7}\scriptsize{$\pm 21.3$} 
& 922.6\scriptsize{$\pm 28.2$} \textcolor{blue}{($\downarrow$ 1.0\%)} & 913.9\scriptsize{$\pm 30.8$} \textcolor{blue}{($\downarrow$ 1.9\%)} 
 \\ 

\textbf{\tabincell{c}{{Cheetah-Vel}}} 
& \textbf{-45.9}\scriptsize{$\pm 4.1$} & -50.2\scriptsize{$\pm 5.2$} \textcolor{blue}{($\downarrow$ 1.9\%)} 
& -55.8\scriptsize{$\pm 2.3$} \textcolor{blue}{($\downarrow$ 4.4\%)} & \textbf{-51.3}\scriptsize{$\pm 4.9$} 
& -125.6\scriptsize{$\pm 7.5$} \textcolor{blue}{($\downarrow$ 33.2\%)} & -208.4\scriptsize{$\pm 1.9$} \textcolor{blue}{($\downarrow$ 76.1\%)} 
 \\ 
 
\midrule

\textbf{\tabincell{c}{{Walker-Param}}} 
& \textbf{368.3}\scriptsize{$\pm 30.6$} & {357.9}\scriptsize{$\pm 33.7$} \textcolor{blue}{($\downarrow$ 2.8\%)} 
& 341.6\scriptsize{$\pm 38.4$} \textcolor{blue}{($\downarrow$ 7.2\%)} & \textbf{287.7}\scriptsize{$\pm 32.1$} 
& 200.1\scriptsize{$\pm 26.3$} \textcolor{blue}{($\downarrow$ 30.4\%)} & 64.7\scriptsize{$\pm 8.1$} \textcolor{blue}{($\downarrow$ 77.5\%)}  
 
\\ 

\textbf{\tabincell{c}{{Hopper-Param}}} 
& \textbf{356.4}\scriptsize{$\pm 16.9$} & 337.0\scriptsize{$\pm 21.2$} \textcolor{blue}{($\downarrow$ 5.4\%)} 
& 319.6\scriptsize{$\pm 14.2$} \textcolor{blue}{($\downarrow$ 10.3\%)} & \textbf{265.2}\scriptsize{$\pm 37.1 $} 
& 159.6\scriptsize{$\pm 35.7$} \textcolor{blue}{($\downarrow$ 39.8\%)} & 82.6\scriptsize{$\pm 15.3$} \textcolor{blue}{($\downarrow$ 68.9\%)}  

\\ \bottomrule 
\end{tabular}
\end{small}
\end{table*}

\subsection{The Robustness of Warm-start Data Quality }
\label{sec: warm-start data quality}
Benefiting from the context encoder and the manner of injecting the explicit context as a label into the diffusion model to conduct the conditional generation, \alg is robust to the quality of warm-start data, similar to traditional context-based methods like CORRO.
\pdt is sensitive to the quality of prompt and the performance can drop a lot with the middle or random prompt, also mentioned in the original paper~\citep{xu2022prompting}. We conduct a more detailed experiment to investigate the robustness of two algorithms.

The results in \cref{table:robustness} show that when the quality of prompt data is not high enough, the performance of \pdt will drop by a large extent except for Cheetah-Dir. This environment contains just two tasks forward and backward, both concluded in the training set and testing set, potentially decreasing the reliance on expert warm-start data. 
The performance of \alg may also experience a slight drop, but still superior to \pdt. The slight drop may be caused by the distribution shift exhibited by poor quality warm-start data during meta-testing and expert data during the pre-trained stage of context encoder, resulting in the inferred context being less accurate. 
For \pdt, the prompt as the prefix to guide the subsequent sequence generation should contain enough valuable knowledge about how to solve the current task, not just information about what the current task is.
But \alg has no strict demand with the quality of warm-start data and even can be rollout with any arbitrary policy. 
The role of warm-start data is just to provide the task-oriented information to the context encoder can infer the task context as the label and then be injected in the conditional denoising process to generate the desired trajectories for planning to fast adaption.

\subsection{The Flexibility in Context Representation Method}
\label{sec:representation}

The generalization ability of \alg arises from capturing task information as context to guide the diffusion model to conditional generation. We argue that our framework can flexibly integrate different task representation algorithm, and the improvement of context accuracy can enhance the generalization performance. We conduct experiments to investigate the effect of different context representations on the few-shot generalization capability of \alg.

\begin{table}[htb]
    \centering
    \vspace{-8pt}
    \caption{The comparisons of the influences of different context representation methods on generalization ability to unseen task.}
    \label{table:representation abla}
    \small
    \scalebox{0.8}{
    \begin{tabular}{cccccc}
    \toprule
    \textbf{Environment}    &\textbf{CORRO}      & \textbf{Ours}        & \textbf{Ours+CORRO} &\textbf{Ours+GT}        \\
    \midrule
    \textbf{Point-Robot}      &-5.59\scriptsize{$\pm 0.57$}       &{-4.48}\scriptsize{$\pm 0.28$}          & {-4.43}\scriptsize{$\pm 0.26$} & \textbf{-4.02}\scriptsize{$\pm 0.13$}          \\
    \textbf{Ant-Dir}       &193.3\scriptsize{$\pm 32.1$}       & {247.7}\scriptsize{$\pm 16.8$}         & {251.3}\scriptsize{$\pm 17.2$}  &\textbf{282.9}\scriptsize{$\pm 13.6$}       \\
    \textbf{Cheetah-Dir}      &823.5\scriptsize{$\pm 37.0$}   & {936.2}\scriptsize{$\pm 17.9$}          & {936.9}\scriptsize{$\pm 18.1$}  &\textbf{939.7}\scriptsize{$\pm 15.7$}      \\
    \textbf{Cheetah-Vel}    &-56.2\scriptsize{$\pm 9.4$}     & {-45.9}\scriptsize{$\pm 4.1$}          & {-44.6}\scriptsize{$\pm 3.9$}  &\textbf{-41.1}\scriptsize{$\pm 3.2$}     \\
    \midrule
    \textbf{Walker-Param}      & 300.5\scriptsize{$\pm 34.2$}        &{368.3}\scriptsize{$\pm 30.6$}           &  {377.0}\scriptsize{$\pm 29.6$} & \textbf{394.1}\scriptsize{$\pm 17.5$}          \\
    \textbf{Hopper-Param}     & 289.3\scriptsize{$\pm 24.7 $}         &{356.4}\scriptsize{$\pm 16.9$}           &  {361.3}\scriptsize{$\pm 19.2$} & \textbf{382.5}\scriptsize{$\pm 12.0$}       \\

    \bottomrule
    \end{tabular}
    }

\end{table}

To this end, we borrowed the representation module of CORRO and integrated it into \alg, shown as Ours+CORRO in \cref{table:representation abla}, resulting in a slight improvement. 
The performance gain demonstrates that the powerful generalization ability of \alg is not achieved by improving context representation capability. The simple representation method we design is not better than the fine-grained representation trained in contrastive learning manner used in CORRO. 
Considering the combination of CORRO representation with \alg can earn a large performance gain than the original conditioned policy manner, conditional sequence modeling shows great potential as a promising paradigm for generalization tasks.

Although we do not seek improvement in generalization performance through a more complicated context representation design in this paper, the incorporation of a more accurate context representation method is always encouraged. The significant improvement in incorporating ground truth parameters of tasks as context into \alg demonstrates that there is still rich room for improvement in the integration of context method, shown as Ours+GT.

\subsection{Ablation Study}

\subsubsection{The Effect of Dual-guide}
\label{abla:dual}

For meta-testing for unseen tasks, the real trajectory rollout with actions in generated trajectory often deviates greatly from the expected trajectory, especially when meeting with a dynamics change environment.
Here we conduct a detailed ablation study to demonstrate the importance of dual-guide for meta-RL setting and gain performance improvements of different relative scaling coefficients between reward guide and dynamics guide in all environments. The visualization in Hopper-Param is shown in \cref{fig:pose_vis} and the results are illustrated in \cref{table:dual guide}.
The utilization of dual-guide can greatly enhance the feasibility and also encourage the high value of generated trajectories when the tasks shift dramatically. In relatively simple environments such as Point-Robot or environments with limited task numbers such as Cheetah-Dir,  overly large dynamics guides can cause diffusion models to generate trajectories that are too conservative and lack a high value to guide. We also tried to omit the value guide and solely utilize the dynamics guide, and found that it yielded relatively poor performance for the same reason.

\begin{table}[htb]
\vspace{-12pt}
\caption{Ablation of dual-guide and relative scaling coefficient. }
\label{table:dual guide}
\centering
\scalebox{0.8}{
\begin{small}
\begin{tabular}{ccccc}
\toprule
\textbf{Environment}  &$\pmb{\lambda = 0}  $ &$\pmb{\lambda = 0.5} $ &$\pmb{\lambda= 1}  $ &$\pmb{\lambda= 2}  $ \\ \midrule

\textbf{\tabincell{c}{{Point-Robot}}}
   & -4.57\scriptsize{$\pm 0.33$} & \textbf{-4.48}\scriptsize{$\pm 0.28$} & -4.74\scriptsize{$\pm 0.26$} & -4.89\scriptsize{$\pm 0.27$} \\

\textbf{\tabincell{c}{{Ant-Dir}}} 
  & 210.3\scriptsize{$\pm 10.7$} & 214.6\scriptsize{$\pm 19.8$} & \textbf{247.7}\scriptsize{$\pm 16.8$} & 238.3\scriptsize{$\pm 18.1$} \\ 

\textbf{\tabincell{c}{{Cheetah-Dir}}} 
  & 924.2\scriptsize{$\pm 19.6$} & \textbf{936.2}\scriptsize{$\pm 17.9$} & 929.7\scriptsize{$\pm 15.1$} & 916.0\scriptsize{$\pm 19.8$}\\ 

\textbf{\tabincell{c}{{Cheetah-Vel}}} 
  & -52.9\scriptsize{$\pm 4.72$} & -49.9\scriptsize{$\pm 2.95$} & \textbf{-45.9}\scriptsize{$\pm 4.1$} & -48.6\scriptsize{$\pm 3.75$}\\ 
\midrule

\textbf{\tabincell{c}{{Walker-Param}}} 
  & 326.5\scriptsize{$\pm 24.9$} & 330.6 \scriptsize{$\pm 23.4$}& 347.2 \scriptsize{$\pm 19.3$}  & \textbf{368.3}\scriptsize{$\pm 30.6$} \\ 

\textbf{\tabincell{c}{{Hopper-Param}}} 
  & 293.3\scriptsize{$\pm 13.8 $} & 
307.2\scriptsize{$\pm 18.6 $}&
328.1\scriptsize{$\pm 16.7 $} &
\textbf{356.8}\scriptsize{$\pm 16.9 $}
\\ \bottomrule 
\vspace{-15pt}
\end{tabular}
\end{small}}
\end{table}

\subsubsection{The comparisons of Generative Models}
\label{sec:cvae}
To investigate the importance of the conditional diffusion model in \alg, we substitute the conditioned diffusion model to conditioned VAE as the same role of trajectory generator to guide the planning across tasks, named as CVAE-Planner. For fairness, the length of generated trajectories and the planning procedure with samples keep the same with \alg.   The results of the experiment are shown in \cref{table:generative model}, demonstrating that the fitting capability of CVAE is significantly inferior to the conditional diffusion model, struggling to generate reasonable trajectories for unseen tasks. Moreover, compared to the end-to-end generative paradigm of CVAE, \alg can fully utilize the gradient from dual-guide 
during the step-wise iterative denoising process.  Additionally, we also trained an unconditional diffusion model over mixed expert data on all the training tasks, named as UDiffuser. UDiffuser, which is the same as vanilla Diffuser in \citep{janner2022diffuser}, struggles to model such a diverse distribution of data and fails to generate the desired trajectories for specific tasks for the lack of ability to infer what the testing task is. 

\begin{table}[t]
    \centering
    \vspace{-6pt}
    \caption{The comparisons between different generative models on generalization ability to unseen task.}
     \label{table:generative model}
     \scriptsize
    \begin{tabular}{ccccc}
    \toprule
    \textbf{Environment}          & \textbf{CVAE-Planner}         &\textbf{UDiffuser} & \textbf{\alg}         \\
    \midrule
    \textbf{Point-Robot}             & -6.11\scriptsize{$\pm 0.71$}          & -7.48\scriptsize{$\pm 0.89$} &\textbf{-4.48}\scriptsize{$\pm 0.28$}          \\
    \textbf{Ant-Dir}              & 130.6\scriptsize{$\pm 41.3$}         & 53.2\scriptsize{$\pm 51.1$}  &\textbf{247.7}\scriptsize{$\pm 16.8$}           \\
    \textbf{Cheetah-Dir}         & 759.4\scriptsize{$\pm 47.2$}          & 614.7\scriptsize{$\pm 63.5$}  &\textbf{936.2}\scriptsize{$\pm 17.9$}       \\
    \textbf{Cheetah-Vel}         & -87.1\scriptsize{$\pm 13.1$}          & -121.7\scriptsize{$\pm 36.2$}  &\textbf{-45.9}\scriptsize{$\pm 4.1$}        \\
    \midrule
    \textbf{Walker-Param}              &187.9\scriptsize{$\pm 49.7$}           &  135.2\scriptsize{$\pm 78.1$} & \textbf{368.3}\scriptsize{$\pm 30.6$}          \\
    \textbf{Hopper-Param}              &157.3\scriptsize{$\pm 54.5$}           &  103.5\scriptsize{$\pm 64.0$} &\textbf{356.4}\scriptsize{$\pm 16.9$}       \\

    \bottomrule
    \end{tabular}
    \vspace{-10pt}
\end{table}









\subsubsection{The distribution shift of data quality}
\label{abla: context data quality}

The data distribution shift in meta-RL stems from the warm-start data and training data quality, which may potentially cause reward or transition shift and the inaccurate guidance from the dual guide during meta-test phase. 
The distribution shift caused by warm-start data has already been studied in \cref{sec: warm-start data quality}, we now take further investigation into the distribution shift of the training data.
Specifically, we replaced the expert dataset used for training the context encoder with mixed data and random data and the training and sampling parts of the conditioned diffusion model remain unchanged. The results in \cref{table:abla on train data quality} show that a completely random dataset performs the worst, while the performance of a dataset that mixes random and expert data surpasses that of the expert dataset in most environments. 
\begin{table}[htbp]
    \centering
    \vspace{-10pt}
    \caption{The comparisons of context encoders trained on datasets of different quality.}
     \label{table:abla on train data quality}
     \scriptsize
    \begin{tabular}{ccccc}
    \toprule
    \textbf{Environment}          & \textbf{Random Dataset}         &\textbf{Mixed Dataset} & \textbf{Expert Dataset}         \\
    \midrule
    \textbf{Point-Robot} & -4.51\scriptsize {$\pm $ }0.25 & -4.49\scriptsize {$\pm $ }0.26 & \textbf{-4.48}\scriptsize {$\pm $ }0.28 \\
    \textbf{Ant-Dir}             & 240.2\scriptsize {$\pm $ }17.1 & \textbf{258.4}\scriptsize {$\pm $ }19.3 & 247.7\scriptsize {$\pm $ }16.8 \\
    \textbf{Cheetah-Dir}        & 936.3\scriptsize {$\pm $ }17.2 & \textbf{936.4}\scriptsize {$\pm $ }17.6 & 936.2\scriptsize {$\pm $ }17.9 \\
    \textbf{Cheetah-Vel}       & -46.8\scriptsize {$\pm $ }4.5  & \textbf{-43.4}\scriptsize {$\pm $ }4.2  & -45.9\scriptsize {$\pm $ }4.1  \\
    \midrule
    \textbf{Walker-Param}      & 359.4\scriptsize {$\pm $ }33.0 & \textbf{381.5}\scriptsize {$\pm $ }28.2 & 368.3\scriptsize {$\pm $ }30.6 \\
    \textbf{Hopper-Param}             & 341.1\scriptsize {$\pm $ }17.3 & \textbf{375.2}\scriptsize {$\pm $ }18.4 & 356.4\scriptsize {$\pm $ }16.9 \\

    \bottomrule
    \end{tabular}
    \vspace{-10pt}
\end{table}

The potential reason may be the diffusion model performs $M$ denoising steps to transform noise into a desired trajectory. In the early denoising steps, the trajectory may be closer to noise or similar to the trajectories from random datasets. The reward guide and dynamics guide for such trajectories with low quality need to have seen this poor distribution during the pretrained phase to provide a more accurate guide than the dual guide only trained on the expert dataset. However, in the later stages of denoising, the trajectory can be improved toward high quality and be more similar to the expert dataset. Therefore, the dual guide trained on a completely random dataset may also be challenging to guide. The differences are not significant in the remaining environments, which may be because the state space of these environments is relatively small, and distribution shift is not a very important factor. Training the context encoder on datasets with a more diverse distribution can provide accurate guidance for the whole denoising process of trajectories gradually denoising from low-quality noise to high-quality desired trajectories.

\subsubsection{The Effect of Context Drop Probability}
\label{abla:drop}
The proper context drop probability can balance off the diversity and the relevance of the conditional label of generated samples~\citep{ho2022classifier}.
We conduct an ablation study with the aim to investigate the effect of context drop probability in the training of the conditional diffusion model. When $\beta$ reaches 1, \alg devolves into the unconditional version previously mentioned as UDiffuser in \cref{table:generative model}.
The results in \cref{table:context drop} show that removing conditional context with a proper probability can improve generalization ability, but the best probability differs from environments. 
One possible explanation for this could be the varying levels of information sharing among tasks in different environments. 
Complex or diverse environments may have higher requirements for conditional generation.

\begin{table}[htb]
\vspace{-8pt}
\caption{Ablation of context drop probability.}
\label{table:context drop}
\centering
\begin{small}
\scalebox{0.8}{
\begin{tabular}{cccccc}
\toprule
\textbf{Environment}  & $\pmb{\beta= 0}$  & $\pmb{\beta = 0.1} $  &$\pmb{\beta= 0.2} $  &$\pmb{\beta  = 0.3} $  \\ \midrule

\textbf{\tabincell{c}{{Point-Robot}}}
  & -4.86\scriptsize{$\pm 0.22$} & -4.61\scriptsize{$\pm 0.34$} & -4.71\scriptsize{$\pm 0.30$} & \textbf{-4.48}\scriptsize{$\pm 0.28$}\\

\textbf{\tabincell{c}{{Ant-Dir}}} 
  & 234.9\scriptsize{$\pm 12.8$} & 241.3\scriptsize{$\pm 16.6$} & \textbf{247.7}\scriptsize{$\pm 16.8$}  & 224.8\scriptsize{$\pm 25.3$} \\ 

\textbf{\tabincell{c}{{Cheetah-Dir}}} 
   &\textbf{936.2}\scriptsize{$\pm 17.9$} & 915.4\scriptsize{$\pm 19.8$} & 909.6\scriptsize{$\pm 23.1$} & 873.8\scriptsize{$\pm 28.6$}\\

\textbf{\tabincell{c}{{Cheetah-Vel}}} 
  & -48.3\scriptsize{$\pm 2.7$} & -49.8\scriptsize{$\pm 3.5$} & -47.4\scriptsize{$\pm 3.7$} & \textbf{-45.9}\scriptsize{$\pm 4.1$} \\ 
\midrule

\textbf{\tabincell{c}{{Walker-Param}}} 
  & 346.5\scriptsize{$\pm 31.4$} & 349.6 \scriptsize{$\pm 35.7$} & \textbf{368.3}\scriptsize{$\pm 30.6$}  & 357.8 \scriptsize{$\pm 29.0$}  \\ 

\textbf{\tabincell{c}{{Hopper-Param}}} 
  & 347.1\scriptsize{$\pm 17.3 $}&
\textbf{356.8}\scriptsize{$\pm 16.9 $}&
336.9\scriptsize{$\pm 12.4 $}    & 
343.3\scriptsize{$\pm 18.6 $}
\\ \bottomrule 

\end{tabular}
}
\end{small}
\end{table}

\subsubsection{The comparisons of denoising steps}


Additionally, we also compared the effects of different denoising steps on trajectory generation.
The experimental results in \cref{table:abla on k} show that relatively longer denoising steps can better denoise desired trajectories from noise, which can slightly improve the quality of generated trajectories. 
Increasing denoising steps provide more chances for the dual guide to precisely manipulate the direction and intensity of denoising, further emphasizing the effectiveness of the dual guide.
Overall, MetaDiffuser is relatively robust to the choice of denoising steps $k$, and its performance still outperforms all baselines.
Due to the fact that more denoising steps mean longer generation time, 
we can consider replacing DDPM~\citep{ho2020denoising} used in this paper with DDIM~\citep{song2020denoising} or DPM solver~\citep{lu2022dpm} for reducing the number of denoising steps to meet the requirement of real-time control.

\begin{table}[t]
    \centering
    \vspace{-6pt}
    \caption{The comparisons between different denoising steps $k$.}
     \label{table:abla on k}
     \scriptsize
    \begin{tabular}{ccccc}
    \toprule
    \textbf{Environment}          & $\pmb{k=20}$        &$\pmb{k=50}$  & $\pmb{k=100}$          \\
    \midrule
    \textbf{Point-Robot}  & \textbf{-4.41}\scriptsize {$\pm $ }0.30 & -4.52\scriptsize {$\pm $ }0.31 & -4.48\scriptsize {$\pm $ }0.28 \\
    \textbf{Ant-Dir}   & 243.1\scriptsize {$\pm $ }15.1 & \textbf{248.0}\scriptsize {$\pm $ }17.2 & 247.7\scriptsize {$\pm $ }16.8 \\
    \textbf{Cheetah-Dir}    & 927.2\scriptsize {$\pm $ }21.4 & 935.2\scriptsize {$\pm $ }16.4 & \textbf{936.2}\scriptsize {$\pm $ }17.9 \\
    \textbf{Cheetah-Vel}        & -47.6\scriptsize {$\pm $ }5.2  & \textbf{-45.4}\scriptsize {$\pm $ }4.3  & -45.9\scriptsize {$\pm $ }4.1  \\
    \midrule
    \textbf{Walker-Param}               & 360.3\scriptsize {$\pm $ }28.7 & 364.8\scriptsize {$\pm $ }31.3 & \textbf{368.3}\scriptsize {$\pm $ }30.6 \\
    \textbf{Hopper-Param}               & 351.5\scriptsize {$\pm $ }15.0 & 356.4\scriptsize {$\pm $ }16.9 & \textbf{360.2}\scriptsize {$\pm $ }16.2      \\

    \bottomrule
    \end{tabular}
    \vspace{-18pt}
\end{table}

\section{Conclusion}
We propose \alg, a novel framework for offline meta-RL, which leverages the diffusion model to conduct conditional trajectory generation to achieve the generalization ability across unseen tasks.
By combining the context representation module with a task-oriented conditional diffusion model to generate the desired trajectories for unseen tasks, \alg demonstrates that the conditional diffusion model can be a promising backbone for offline meta-RL. Moreover, we design the dual-guide to improve the quality of generated trajectories in the sampling process, ensuring dynamics transition consistency with the real world while encouraging the generated trajectories to achieve high returns. 
The experiments on various benchmarks empirically show that \alg much better generalizes to unseen tasks than prior methods, while also enjoying both the flexibility to incorporate with other task representation methods and the robustness to the quality of collected warm-start data at the testing task.

\textbf{Limitation.}
Although \alg enjoys the robustness of warm-start data, the framework still faces the limitation of the need for expert training data in meta-training phase, which is the common dilemma in offline meta-RL. Besides, \alg has not been evaluated in real robots and the requirements of real-time control may be challenging. 

\textbf{Future Work.}
Further improving the speed of real-time trajectory generation in planning and supporting high-dimensional image inputs are directions for future work. Additionally, the combination of a large language model (LLM) with the reasoning ability in complex control tasks with \alg is an interesting research direction.

\section*{Acknowledgements}
This work is supported by the National Key R\&D Program of China (Grant No. 2022ZD0116402), the National Natural Science Foundation of China (Grant No. 62106172), and the Natural Science Foundation of Tianjin (No. 22JCQNJC00250).

\nocite{langley00}

\bibliography{example_paper}
\bibliographystyle{icml2023}

\newpage
\appendix
\onecolumn

\section{Classifier-Free and Classifier-Guided Diffusion Model}
\label{appedix:classifier}

In this section, we introduce the details of the theoretical analysis of the conditional diffusion model. 
The equivalence between diffusion models and score-matching~\citep{song2020denoising}, which shows $\epsilon_\theta(\boldsymbol{x}_{k}, k) \propto \nabla_{\boldsymbol{x}_{k}}\log p(\boldsymbol{x}_{k})$, naturally leads to two kinds of methods for conditioning: classifier-guided~\citep{nichol2021improved} and classifier-free~\citep{ho2022classifier}.The classifier-guided improves sample quality while reducing diversity in conditional diffusion models using gradients from a pre-trained classifier $p_{\phi}(\boldsymbol{y}|\boldsymbol{x}_{k})$ during sampling. The classifier-free is an alternate technique that avoids this pre-trained classifier by instead jointly training a single diffusion model on conditional $\epsilon_{\theta}(\boldsymbol{x}_{k}, \boldsymbol{y}, k)$ and unconditional $\epsilon_{\theta}(\boldsymbol{x}_{k}, k)$ noise model via randomly dropping conditional label $\boldsymbol{y}$.

\subsection{Classifier-Guided Fashion}
First, let us start with classifier-guided fashion. The initial conditional distribution that conditions on the respective label $\boldsymbol{y}$ can be formulated by Bayes rule as:
\begin{equation}
\label{eq:bayes}
    p\left(\boldsymbol{x}_{k-1} \mid \boldsymbol{y}\right)=\frac{p\left(\boldsymbol{x}_{k-1}\right) p\left(\boldsymbol{y} \mid \boldsymbol{x}_{k-1}\right)}{p(\boldsymbol{y})}
\end{equation}

where $k$ denotes the timestep of the denoising process.The most important advantage of the classifier guide is that it can reuse previously trained unconditional generation models $p\left(\boldsymbol{x}_{k-1} \mid \boldsymbol{x}_k\right)$. By training an additional classifier $p\left(\boldsymbol{y} \mid \boldsymbol{x}_{k-1}\right)$ on the generated samples, its evaluation about the generated samples can be used as a gradient to guide the noise model during denoising process. Now we consider how to incorporate the unconditional diffusion model into the conditional diffusion model. With the additional condition of the current noisy sample $\boldsymbol{x}_k$, Eq.\eqref{eq:bayes} can be rewritten as:
\begin{equation}
\label{rewritten}
\begin{aligned}
    p\left(\boldsymbol{x}_{k-1} \mid \boldsymbol{x}_k, \boldsymbol{y}\right)
    &=\frac{p\left(\boldsymbol{x}_{k-1} \mid \boldsymbol{x}_k\right) p\left(\boldsymbol{y} \mid \boldsymbol{x}_{k-1}, \boldsymbol{x}_k\right)}{p\left(\boldsymbol{y} \mid \boldsymbol{x}_k\right)}\\
    & =\frac{p\left(\boldsymbol{x}_{k-1} \mid \boldsymbol{x}_k\right) p\left(\boldsymbol{y} \mid \boldsymbol{x}_{k-1}\right)}{p\left(\boldsymbol{y} \mid \boldsymbol{x}_k\right)}\\
    &=p\left(\boldsymbol{x}_{t-1} \mid \boldsymbol{x}_k\right) e^{\log p\left(\boldsymbol{y} \mid \boldsymbol{x}_{k-1}\right)-\log p\left(\boldsymbol{y} \mid \boldsymbol{x}_k\right)} \\
\end{aligned}
\end{equation}
It is worth noting that $\boldsymbol{x}_k$ are obtained by diffusing over $\boldsymbol{x}_{k-1}$ with noise, which is not helpful for classifier evaluation. So we can assume $p\left(\boldsymbol{y} \mid \boldsymbol{x}_{k-1}, \boldsymbol{x}_k\right)=p\left(\boldsymbol{y} \mid \boldsymbol{x}_{k-1}\right)$. When the diffusing steps are large enough, the difference between  $\boldsymbol{x}_k$ and $\boldsymbol{x}_{k-1}$ is tiny. So we apply Taylor's Formula to the exponent term in Eq.\eqref{rewritten} and can get: 
\begin{equation}
\label{talyer}
    \log p\left(\boldsymbol{y} \mid \boldsymbol{x}_{k-1}\right)-\log p\left(\boldsymbol{y} \mid \boldsymbol{x}_k\right) \approx\left(\boldsymbol{x}_{k-1}-\boldsymbol{x}_k\right) \cdot \nabla_{\boldsymbol{x}_k} \log p\left(\boldsymbol{y} \mid \boldsymbol{x}_k\right)
\end{equation}
For another term, $p\left(\boldsymbol{x}_{k-1} \mid \boldsymbol{x}_k\right)$, also as the unconditional diffusion model, can be rewritten in form of distribution:
\begin{equation}
\label{unconditional}
    p\left(\boldsymbol{x}_{k-1} \mid \boldsymbol{x}_k\right)=\mathcal{N}\left(\boldsymbol{x}_{k-1} ; \boldsymbol{\mu}\left(\boldsymbol{x}_k\right), \sigma_k^2 \boldsymbol{I}\right) \propto e^{-\left\|\boldsymbol{x}_{k-1}-\boldsymbol{\mu}\left(\boldsymbol{x}_k\right)\right\|^2 / 2 \sigma_k^2}
\end{equation}
where the mean $\mu$ and variance $\sigma$ denotes the mean and variance of the Gaussian distribution respectively.
With the above Eq.\eqref{talyer} and Eq.\eqref{unconditional}, we have:
\begin{equation}
    \begin{aligned}
p\left(\boldsymbol{x}_{k-1} \mid \boldsymbol{x}_k, \boldsymbol{y}\right) & \propto e^{-\left\|\boldsymbol{x}_{k-1}-\boldsymbol{\mu}\left(\boldsymbol{x}_k\right)\right\|^2 / 2 \sigma_k^2+\left(\boldsymbol{x}_{k-1}-\boldsymbol{x}_k\right) \cdot \nabla_{\boldsymbol{x}_k} \log p\left(\boldsymbol{y} \mid \boldsymbol{x}_k\right)} \\
& \propto e^{\left.-\| \boldsymbol{x}_{k-1}-\boldsymbol{\mu}\left(\boldsymbol{x}_k\right)-\sigma_k^2 \nabla_{\boldsymbol{x}_k} \log p\left(\boldsymbol{y} \mid \boldsymbol{x}_k\right)\right) \|^2 / 2 \sigma_k^2}
\end{aligned}
\end{equation}
Based on this proportional property, now we can obtain the classifier-guided conditioned diffusion model $p\left(\boldsymbol{x}_{k-1} \mid \boldsymbol{x}_k, \boldsymbol{y}\right)$:
\begin{equation}
\begin{aligned}
        \left.p\left(\boldsymbol{x}_{k-1} \mid \boldsymbol{x}_k, \boldsymbol{y}\right) \approx \mathcal{N}\left(\boldsymbol{x}_{k-1} ; \boldsymbol{\mu}\left(\boldsymbol{x}_k\right)+\sigma_k^2 \nabla_{\boldsymbol{x}_k} \log p\left(\boldsymbol{y} \mid \boldsymbol{x}_k\right)\right), \sigma_k^2 \boldsymbol{I}\right) \\
        \Rightarrow \boldsymbol{x}_{k-1}=\boldsymbol{\mu}\left(\boldsymbol{x}_k\right)+\sigma_k^2 \nabla_{x_k} \log p\left(\boldsymbol{y} \mid x_k\right)+\sigma_k \boldsymbol{\varepsilon}, \quad \boldsymbol{\varepsilon} \sim \mathcal{N}(\mathbf{0}, \boldsymbol{I})
\end{aligned}
\end{equation}
This can also be formulated as the denoising version with noise model $\hat{\epsilon}\left(\boldsymbol{x}_k\right)$:
\begin{equation}
\hat{\epsilon}\left(\boldsymbol{x}_k\right):=\epsilon_\theta\left(\boldsymbol{x}_k\right)-\sqrt{1-\bar{\alpha}_k} \nabla_{\boldsymbol{x}_k} \log p_\phi\left(\boldsymbol{y} \mid \boldsymbol{x}_k\right)
\end{equation}

\subsection{Classifier-Free Fashion}
Classifier-free fashion is relatively easy to understand compared with classifier-guided fashion. This manner does not separately train a classifier but modifies the original training setup to learn both a conditional $\epsilon_{\theta}(\boldsymbol{x}_{k}, \boldsymbol{y}, k)$ and an unconditional $\epsilon_{\theta}(\boldsymbol{x}_{k}, k)$ model for the noise.

Without extra classifier $p\left(\boldsymbol{y} \mid \boldsymbol{x}_{k-1}\right)$ to reuse, we can directly define the conditional data distribution $p\left(\boldsymbol{x}_{k-1} \mid \boldsymbol{x}_k, \boldsymbol{y}\right)$ with  conditional label $\boldsymbol{y}$:
\begin{equation}
    \begin{aligned}
p\left(\boldsymbol{x}_{k-1} \mid \boldsymbol{x}_k, \boldsymbol{y}\right)=\mathcal{N}\left(\boldsymbol{x}_{k-1} ; \boldsymbol{\mu}\left(\boldsymbol{x}_k, \boldsymbol{y}\right), \sigma_k^2 \boldsymbol{I}\right) \\
\boldsymbol{\mu}\left(\boldsymbol{x}_k, \boldsymbol{y}\right)=\frac{1}{\alpha_k}\left(\boldsymbol{x}_k-\frac{\beta_k^2}{\bar{\beta}_k} \boldsymbol{\epsilon}_{\boldsymbol{\theta}}\left(\boldsymbol{x}_k, \boldsymbol{y}, k\right)\right)
\end{aligned}
\vspace{-2pt}
\end{equation}
where $\alpha_k, \bar{\beta}_k  \in \mathbb{R}$ are carefully chosen for the variance schedule during the diffuse process. The noise model $\boldsymbol{\epsilon}_{\boldsymbol{\theta}}$ in the denoising process can be trained by minimizing the reconstruction error about noise, following as:
\begin{equation}
    \mathbb{E}_{\boldsymbol{x}_0, \boldsymbol{y} \sim \tilde{p}\left(\boldsymbol{x}_0, \boldsymbol{y}\right), \boldsymbol{\varepsilon} \sim \mathcal{N}(\mathbf{0}, \boldsymbol{I})}\left[\left\|\boldsymbol{\varepsilon}-\boldsymbol{\epsilon}_{\boldsymbol{\theta}}\left(\bar{\alpha}_k \boldsymbol{x}_0+\bar{\beta}_k \boldsymbol{\varepsilon}, \boldsymbol{y}, k\right)\right\|^2\right]
\end{equation}
It is worth noting that the conditional label is randomly dropped with probability $\beta$, following the classifier-free conditioned diffusion model~\citep{ho2022classifier}. The potential reason is the proper context drop probability can balance off the diversity and the relevance of the conditional label of generated samples. At testing-time, the perturbed noise is generated as:
\begin{equation}
    \hat{\epsilon}_\theta\left(\boldsymbol{x}_{k} \mid \boldsymbol{y}\right)=\epsilon_{\theta}(\boldsymbol{x}_{k}, k) + \omega(\epsilon_{\theta}(\boldsymbol{x}_{k}, \boldsymbol{y}, k) - \epsilon_{\theta}(\boldsymbol{x}_{k}, k))
\end{equation}
where $\omega$ is referred to as the guidance scaling, similar to classifier-guided fashion.
Setting $\omega=1$ enables classifier-free guidance while increasing $\omega>1$ strengthens the effect of guidance.

\section{Further Introduce of Sequence Modeling Fashion}

Recently, much attention has been focused on the use of large, pre-trained big models on unsupervised datasets to improve results on downstream tasks through fine-tuning. In the field of natural language processing, transformer-based models such as BERT\citep{bert} and GPT-3\citep{brown2020gpt3} have overcome the limitations of RNNs and improved the ability to use long sequence information, resulting in state-of-the-art performance on tasks such as translation and question answering systems. The field of computer vision has also been inspired by these developments, with models like the Swin Transformer\citep{SwinTransformer} and Scaling ViT\citep{scalingVIT} being proposed to address problems as sequence modeling problems.

Traditional offline RL approaches require fitting value functions or computing policy gradients, which are challenging due to limited offline data~\cite{kumar2020conservative,wu2019behavior,kidambi2020morel}. 
Inspired by the exciting progress of large generative models in vision and language tasks, researchers turn to model the trajectories in offline RL datasets through transformer-like structures. 
Recent advances in generative sequence modeling~\cite{chen2021decision, janner2021offline, janner2022planning} provide effective alternatives to conventional RL problems by modeling the joint distribution of sequences of states, actions, rewards, and values.
For example, Decision Transformer~\cite{chen2021decision} casts offline RL as a form of conditional sequence modeling, which allows more efficient and stable learning without the need to train policies via traditional RL algorithms like temporal difference learning~\cite{sutton1988learning}.
Trajectory Transformer~\cite{janner2021offline} is proposed to utilize transformer architecture to model distributions over trajectories, repurposes beam search as a planning algorithm, and shows great flexibility across long-horizon dynamics prediction, imitation learning, goal-conditioned RL, and offline RL. Bootstrapped Transformer~\cite{wang2022bootstrapped} further incorporates the idea of bootstrapping and uses the learned model to self-generate more offline data to further improve sequence model training.  MAT\citep{MAT} introduces sequence modeling into the online MARL setting and demonstrates high data efficiency in transfer.
By treating RL as a sequence modeling problem, it bypasses the need for bootstrapping for long-term credit assignment, avoiding one of the "deadly triad"~\cite{sutton2018reinforcement} challenges in reinforcement learning.

The above methods mostly focus on Transformer-like architecture, and recent works begin to adopt the diffusion model as the backbone of sequence modeling. Diffuser~\citep{janner2022planning} applies a diffusion model as a trajectory generator, which is trained by diffusing over the full trajectory of state-action pairs from the noises.
A separate reward model is trained to predict the cumulative rewards of each trajectory sample, then the gradient guidance from the reward model is injected into the reverse sampling stage. Then the first action in the generated trajectories will be applied to execute in the environment to step into the next state, which repeats in a loop until the terminal. The consequent work Decision Diffuser~\citep{ajay2022conditional} frames offline sequential decision-making as conditional generative modeling based on returns, constraints, and skills to eliminate the complexities in traditional offline RL.

\section{Additional Discussion of Offline Meta-RL}
\label{app:additional discussion}
One of the inherent difficulties of offline RL is the challenge to generalize to unseen tasks. Recent work in offline meta-RL ~\citep{mitchell2021offline,mbml,improved-context,focal} aims to solve this problem by training a meta-policy from multi-task offline datasets that can efficiently adapt to unseen tasks with small amounts of warm-start data.

 The context-based offline meta RL methods pre-train a context encoder to learn task representation from the collected offline data and augment the state-action pair with latent representation to generalize across tasks. 
MACAW \citep{MACAW} adopts the advantage weighting loss based on an optimization-based method and learns the initialization of both the value function and the policy.
FOCAL~\citep{focal} proposes a novel
negative-power distance metric learning method to train the context encoder for task inference, as an end-to-end offline meta-RL algorithm with high efficiency. 
CORRO~\citep{yuan2022robust} proposes a contrastive learning framework for task representations that are robust to the distribution mismatch of behavior policies in training and testing. CORRO demonstrates superior performance than prior context-conditioned policy-based methods.

The context-based methods can infer the current task from the pre-collected offline data as long as it contains enough information about the current task, regardless of the quality of data. But one of the limitations of these methods is the high requirements for the representation ability of the context. 
If the context is not sufficiently accurate, it will negatively impact the generalization performance of policy across tasks, which severely relies on the representation ability of context. Although many recent works~\cite{mbml,borel,improved-context,focal} focused on improving the representation accuracy of context, the methods still rely on TD-learning, which may potentially cause instability in context learning.

\begin{table}[h!]
\vspace{-6pt}

\caption{A comparison on algorithmic properties of existing methods for offline meta-RL.}
\centering
\resizebox{0.7\textwidth}{!}{
    \begin{tabular}{cccc}
		\toprule
		Algorithm &  Robustness & Flexibility & Sequence Modeling  \\
		\midrule
  MACAW~\cite{MACAW} & \Checkmark & \Checkmark & \XSolidBrush \\ 
		FOCAL~\cite{focal} & \Checkmark & \Checkmark & \XSolidBrush \\ 
		CORRO~\cite{yuan2022robust}& \Checkmark & \Checkmark & \XSolidBrush \\ 
		\pdt~\cite{xu2022prompting} & \XSolidBrush & \XSolidBrush & \Checkmark \\
		\midrule
		\alg (Ours) & \textcolor{green}{\Checkmark} & \textcolor{green}{\Checkmark} & \textcolor{green}{\Checkmark} \\
		\bottomrule
    \end{tabular}
}

\label{table:exsiting comparision}
\vspace{-8pt}
\end{table}
\vspace{-2pt}

A more recent work \pdt~\citep{xu2022prompting} turns to solve the generalization problem from the sequence modeling perspective, which joint models state-action trajectories to avoid TD-learning.
This approach does not require a pre-trained context encoder to extract the task representation from the prompt and the contained task-oriented information of the prompt is implicitly learned inside the generative model architecture based on powerful ability transformer-style models. 
The prompt as the prefix to guide the subsequent sequence generation should not only contain rich task information to allow the model to implicitly infer the current task but also has to conclude the expert trajectories as valuable knowledge to utilize. Therefore, the core limitation of the method is the quality of the pre-collected demonstration or prompt has to be high enough to guide the sequence generation or the performance will drop a lot with random or medium prompts. It is worth noting that we can not easily obtain expert trajectories on all tasks, which is challenging and computationally high in many scenarios.
A more clear comparison of algorithmic properties of existing methods for offline meta-RL is shown in \cref{table:exsiting comparision}.

\newpage
\section{Pesudocodes of Framework}
\label{app:pesudocodes}
\begin{algorithm}[h]
\caption{Task-Oriented Conditioned Diffusion Planner for Offline Meta-RL~(\alg)}
\begin{algorithmic}

\STATE \textbf{Input}: Training tasks set $\gT^{train}$ and corresponding offline datasets $\gD^{train}$, testing tasks set $\gT^{test}$ and corresponding warm-start datasets $\gD^{test}$, noise model $\epsilon_\theta$, guidance scale $s$, historical trajectory length $h$, planning horizon $H$, context drop probability $\beta$,  context encoder $E_{\phi}$, reward predictor $R_\psi$, dynamics model $P_w$.
\begin{tcolorbox}[colback=red!5!white,colframe=red!50!black!50!,left=2pt,right=2pt,top=0pt,bottom=1pt,colbacktitle=red!25!white,title=\textbf{\color{black} Pre-training Context Encoder}]
\FOR{each iteration}
\STATE Sample a task $M \sim \gT^{train} $ and corresponding trajectories $\tau^M$ from $\gD^{train}$. 
\STATE Sample several historical transitions $\tau^\gM_t =\{(s_{t+i},a_{t+i},r_{t+i},s_{{t+i+1}})\}_{i=0}^h$  started from random selected $t$
\STATE Predict the context from the historical transitions: $z_t = E_{\phi}(\tau^\gM_t)$
\STATE Update $\phi$, $\psi$, $\omega$ on $\tau^M$ according the following loss: 
\STATE $\mathcal{L}_{\phi,\psi,\omega}= -\mathbb{E}_{\left(s_t, a_t, r_t, s_{t+1}\right) \sim \tau^\gM_t;\gM \sim \gT^{train} } \Big[ \mathbb{E}_{z_t=E_\phi(z_t \mid \tau^\gM_t)} \left[\log P_\omega(\hat{s}_{t+1}|s_t, a_t, z_t)+ \log R_\psi(\hat{r}_t|s_t, a_t, z_t) \right] \Big]$
\ENDFOR
\end{tcolorbox}

\begin{tcolorbox}
[colback=yellow!5!white,colframe=yellow!50!black!50!,left=2pt,right=2pt,top=0pt,bottom=1pt,colbacktitle=yellow!25!white,title=\textbf{\color{black} Training Task-Oriented Diffusion Model}]
\FOR{each iteration}
\STATE Sample trajectories with planning horizon $H$ as mini-batch $\mathcal{B}=\{\tau^\gM ; \gM \sim \gT^{train} \}$ from mixed  $\gD^{train}$.

\STATE Predict the context $z_\tau$ from context encoder for each $\tau$ in $\mathcal{B}$
\STATE Random sample a denosing timestep $k$ and omit the context with probability $\beta$
\STATE Update the task-oriented conditioned diffusion model according the following loss:


\STATE $\mathcal{L}(\theta):=\mathbb{E}_{k, \tau \in \mathcal{D}, p \sim \operatorname{Bern}(\beta)} \left[\left\|\epsilon-\epsilon_\theta\left(\boldsymbol{x}_k\left(\tau\right),(1-p) E_\phi\left(\tau\right)+p\varnothing, k\right)\right\|^2\right]$

\ENDFOR
\end{tcolorbox}

\begin{tcolorbox}
[colback=teal!5!white,colframe=teal!50!black!50!,left=2pt,right=2pt,top=0pt,bottom=1pt, colbacktitle=teal!25!white,title=\textbf{\color{black} Testing on Unseen Tasks}]
\FOR{each iteration}
\STATE Sample a task $M \sim \gT^{test} $ and corresponding warm-start data $\tau^M$ from $\gD^{test}$. 
\STATE Random sample a sub-segment of trajectories from $\tau^M$
\STATE Infer the context for the current test task from context encoder capturing task-oriented information 
\WHILE {not done}
\STATE Observe state $s$ ; Initialize $\boldsymbol{x}_T(\tau) \sim \mathcal{N}(0, \alpha I)$
\FOR{$t=T\dots1$}
\STATE $    \hat{\epsilon} \coloneqq  \underbrace{\omega \epsilon_\theta(\boldsymbol{x}_k(\tau), E_\phi(\tau), k) + (1 - \omega)\epsilon_\theta(\boldsymbol{x}_k(\tau), \varnothing, k)}_{\text{classifier-free}}  - \underbrace{\sqrt{1-\bar{\alpha}_t}  \nabla _{\boldsymbol{x}_{k}(\tau)} \Big[\mathcal{J}(\boldsymbol{x}_{k}(\tau)) + \lambda \mathcal{\zeta}(\boldsymbol{x}_{k}(\tau))\Big]}_{\text{classifier-guided}}$
\STATE $\left(\mu_{t-1}, \Sigma_{t-1}\right) \leftarrow \operatorname{Denoise}\left(\boldsymbol{x}_t(\tau), \hat{\epsilon}\right)$
\STATE $\boldsymbol{x}_{t-1} \sim \mathcal{N}\left(\mu_{t-1}, \alpha \Sigma_{t-1}\right)$
\ENDFOR
\STATE Execute first action from $\boldsymbol{x}_0$ as the current action to interact with environments.
\ENDWHILE  
\ENDFOR
\end{tcolorbox}

\end{algorithmic}
\end{algorithm}

\section{The Details of Environments}
\label{app:benchmark}

We adopt a 2D navigation environment Point-Robot and multi-task MuJoCo control tasks to make comparisons, as classical benchmarks commonly used in meta-RL~\citep{mitchell2021offline,mbml,improved-context}. The environments conclude $4$ tasks with reward function changes and $2$ tasks with transition dynamics changes. 

\begin{itemize}
    \item \textbf{Point-Robot} is a 2D navigation environment introduced in \citet{pearl}. Starting from the initial point, the agent should navigate to the goal location. Tasks differ in reward functions, which describe the goal position. The goal positions are uniformly distributed in a square. The maximal episode step is set to $20$.
    \item \textbf{Cheetah-Vel,  Cheetah-Dir, Ant-Dir} are multi-task MuJoCo benchmarks where tasks differ in reward functions. In Cheetah-Vel, the task is specified by the target velocity of
the agent. The distribution of target velocity is $U[0, vmax]$.
In Cheetah-Dir, the goal directions are limited to forward and backward.
In Ant-Dir, the task is specified by the goal direction of
the agent’s motion. The distribution of goal direction is
$U[0, 2\pi]$.The maximal episode step is set to $200$.
    \item \textbf{Walker-Param, Hopper-Param} are multi-task MuJoCo
benchmarks where tasks differ in transition dynamics. For
each task, the physical parameters of body mass, inertia,
damping, and friction are randomized. The agent should
adapt to the varying environment dynamics to accomplish
the task. The maximal episode step is set to $200$. 
\end{itemize}
For all environments except for Cheetah-Dir, $40$ tasks are randomly sampled with different goal velocities, locomotion, directions, or different physical parameters.
we sample $10$ tasks for meta-testing and leave the rest for meta-training.
For Cheetah-Dir, with only two tasks limited in forward and backward, we keep the training set the same as the testing set, both containing the two tasks.

\section{Additional Experiment Results}

\subsection{The Results on MetaWorld}

We additionally choose a more realistic robotic manipulation benchmark Meta-World~\citep{yu2020meta}. 
We have currently added three typical tasks: reach, push, and pick-place for evaluation. The results with 5 runs for each task are as follows:

\begin{table}[htbp]
\caption{The comparisons of the performance of Prompt-DT and \alg  in typical tasks from MetaWorld.} 

\label{table:robustness on metaworld}
\centering
\small
\begin{tabular}{@{}ccccc@{}}
\toprule
\multicolumn{1}{c}{\multirow{2}{*}{\textbf{Environment}}} & \multicolumn{2}{c}{\textbf{Prompt-DT}} & \multicolumn{2}{c}{\textbf{\alg}}  \\ \cmidrule(lr){2-5} 
\multicolumn{1}{c}{}  &  \textit{Random} & \textit{Expert}  &  \textit{Random} & \textit{Expert}\\ \midrule


\textbf{\tabincell{c}{{Reach-v2}}}
& 3686.4 \scriptsize {$\pm $ }208.3 & 4358.5\scriptsize {$\pm $ }234.4 & 4671.3\scriptsize {$\pm $ }113.6 & \textbf{4708.7}\scriptsize {$\pm $ }68.4  \\

\textbf{\tabincell{c}{{Push-v2}}} 
& 3638.1\scriptsize {$\pm $ }203.6  & 4010.2\scriptsize {$\pm $ }140.7 & 4226.3\scriptsize {$\pm $ }160.5 & \textbf{4352.4}\scriptsize {$\pm $ }107.7 \\

\textbf{\tabincell{c}{{Pick-and-Place-v2}}} 
& 3524.1\scriptsize {$\pm $ }337.4 & 3994.6\scriptsize {$\pm $ }290.5 & 4161.3\scriptsize {$\pm $ }243.0 & \textbf{4205.8}\scriptsize {$\pm $ }217.3 \\
\bottomrule 
\vspace{-14pt}
\end{tabular}
\end{table}

The results show that MetaDiffuser still performs better than Prompt-DT in this more challenging and realistic environment. To demonstrate our robustness, we also conducted an experiment on the influence of warm-start data quality, and verify the robustness of MetaDiffuser with the benefit of task representation. 

\subsection{The Effect of Guidance Scaling $\omega$}

\begin{table}[htb]
\vspace{-12pt}
\caption{The comparisons between different guidance scaling $\omega$.}
\label{table: omega}
\centering
\small
\begin{tabular}{cccccc}
\toprule
\textbf{Environment}  & $\pmb{\omega= 1.2}$  & $\pmb{\omega = 1.4} $  &$\pmb{\omega= 1.6} $ &$\pmb{\omega  = 1.8} $  &$\pmb{\omega  = 2.0} $  \\ \midrule

\textbf{\tabincell{c}{{Point-Robot}}}
  & \textbf{-4.48}\scriptsize {$\pm $ }0.28 & -4.51\scriptsize {$\pm $ }0.23 & -4.55\scriptsize {$\pm $ }0.18 & -4.52\scriptsize {$\pm $ }0.31 & -4.49\scriptsize {$\pm $ }0.30 \\

\textbf{\tabincell{c}{{Ant-Dir}}} 
& 244.9\scriptsize {$\pm $ }20.1 & \textbf{247.7}\scriptsize {$\pm $ }16.8 & 243.2\scriptsize {$\pm $ }19.2 & 241.1\scriptsize {$\pm $ }12.6 & 246.0\scriptsize {$\pm $ }16.3 \\

\textbf{\tabincell{c}{{Cheetah-Dir}}} 
   & 935.9\scriptsize {$\pm $ }19.5 & 933.1\scriptsize {$\pm $ }20.2 & 935.7\scriptsize {$\pm $ }18.8 & 932.2\scriptsize {$\pm $ }19.1 & \textbf{936.2}\scriptsize {$\pm $ }17.9 \\

\textbf{\tabincell{c}{{Cheetah-Vel}}} 
  & -45.5\scriptsize {$\pm $ }3.7  & \textbf{-45.3}\scriptsize {$\pm $ }3.9  & -45.9\scriptsize {$\pm $ }4.1  & -46.7\scriptsize {$\pm $ }5.0  & -46.1\scriptsize {$\pm $ }5.2  \\
\midrule

\textbf{\tabincell{c}{{Walker-Param}}} 
   & 367.1\scriptsize {$\pm $ }24.9 & 362.6\scriptsize {$\pm $ }24.5 & 369.1\scriptsize {$\pm $ }27.7 & 368.3\scriptsize {$\pm $ }30.6 & \textbf{375.0}\scriptsize {$\pm $ }29.9 \\

\textbf{\tabincell{c}{{Hopper-Param}}} 
& 355.4\scriptsize {$\pm $ }16.8 & 353.7\scriptsize {$\pm $ }16.2 & \textbf{362.1}\scriptsize {$\pm $ }18.0 & 358.1\scriptsize {$\pm $ }19.2 & 356.4\scriptsize {$\pm $ }16.9
\\ \bottomrule 

\end{tabular}
\end{table}

The scalar $\omega$ denotes the guidance weight in the classifier-free conditioned diffusion model. Setting $\omega = 1$ disables classifier-free guidance while increasing $\omega > 1$ strengthens the effect of guidance. The relationship between performance and $\omega$ does not show a clear trend, and the choice of different $\omega$ values does not have a significant impact. 

\subsection{The Effect of Historical Length $h$}

The longer historical trajectories can bring a slight improvement with the benefit of more task-oriented information contained. In the submitted version, we choose the historical trajectories length $h$ of $4$ in Point-Robot tasks, $10$ in Ant-Dir, Cheetah-vel, and Cheetah-Dir tasks with reward change, $20$ in Hopper-Param and Walker-Param tasks with dynamics change. Note that the maximum length within the Point environment is 20, so we skip this environment for testing longer $h$. In a cheetah-dir environment with only two tasks(forward and backward), 10 transitions are enough for the context encoder to infer the current task and the longer transitions make no more benefit. In a complex environment with dynamics change, longer transitions can conclude more task information to help for quick adaptation.

\begin{table}[htbp]
    \centering
    \vspace{-10pt}
    \caption{The comparisons between different historical length $h$.}
     \label{table:abla on h}
     \small
    \begin{tabular}{ccccc}
    \toprule
    \textbf{Environment}          & $\pmb{h=10}$         &$\pmb{h=20}$ & $\pmb{h=40}$         \\
    \midrule
    \textbf{Ant-Dir}  & 247.7\scriptsize {$\pm $ }16.8 & 248.3\scriptsize {$\pm $ }17.2 & \textbf{249.2}\scriptsize {$\pm $ }15.5 \\
    \textbf{Cheetah-Dir}    & 936.2\scriptsize {$\pm $ }17.9 & \textbf{936.9}\scriptsize {$\pm $ }17.0 & 936.7\scriptsize {$\pm $ }17.6 \\
    \textbf{Cheetah-Vel}       & -45.9\scriptsize {$\pm $ }4.1  & \textbf{-44.6}\scriptsize {$\pm $ }5.2  & -45.7\scriptsize {$\pm $ }3.9  \\
    \midrule
    \textbf{Walker-Param}            & 360.9\scriptsize {$\pm $ }32.3 & 368.3\scriptsize {$\pm $ }30.6 & \textbf{371.9}\scriptsize {$\pm $ }28.6 \\
    \textbf{Hopper-Param}               & 349.4\scriptsize {$\pm $ }18.7 & 356.4\scriptsize {$\pm $ }16.9 & \textbf{362.0}\scriptsize {$\pm $ }15.3     \\

    \bottomrule
    \end{tabular}
    \vspace{-8pt}
\end{table}




\section{Details of Implementations}
\label{app:implementation}
For pre-training expert policy for each task, we borrow the provided scripts in the official code repositories from CORRO\footnote{\scriptsize\url{https://github.com/PKU-AI-Edge/CORRO}}. All baselines are trained with the same offline dataset collected with the pre-trained expert policy. 

For FOCAL and CORRO, we run with the open-source implementations\footnote{\scriptsize\url{https://github.com/LanqingLi1993/FOCAL-ICLR}} for fair comparisons. 
For \pdt, note that the open-source
implementation of \pdt \footnote{\scriptsize\url{https://github.com/mxu34/prompt-dt}} specializes in different hyper-parameter configurations for different environments, such as target rewards and prompt length. In this paper, we conduct three environments, Point-Robot, Walker-Param, and Hopper-Param,
without official hyper-parameter configuration since \pdt does not evaluate them. We select the best target rewards and prompt length for each task from the set of hyper-parameters used in the original paper.
Moreover, the prompts with medium and random quality are not provided in the official code repositories, so we collect demonstrations during the policy pre-training process as prompts. The slight difference is that we strictly use the accumulated rewards of the trajectories as the metric to distinguish the quality of the prompt. \pdt divides the prompt into three qualities based on the training epochs, which may result in the quality of medium prompts approaching that of expert prompts when training converged fast for some tasks, which could bring a confusing conclusion.

For CVAE-Planner, we substitute the conditioned diffusion to conditioned VAE, which serves the same role as a trajectory generator to guide the planning across tasks. The generated trajectories conclude the state and actions, similar to \alg. The planning horizon stays the same with \alg for fairness to investigate the difference between CVAE and conditional diffusion model.

\section{Hyperparameter and Architectural details}
\label{app:hyperparam}
In this section, we describe various architectural and hyperparameter details:
\begin{itemize}[leftmargin=*]

\item We represent the noise model $\epsilon_\theta$ with a temporal U-Net~\citep{janner2022diffuser}, consisting of
a U-Net structure with 6 repeated residual blocks.
Each block consisted of two temporal convolutions,
each followed by group norm~\citep{wu2018groupnorm}, and
a final Mish nonlinearity~\citep{misra2019mish}.

\item We choose the historical trajectories length $h$ of $4$ in Point-Robot tasks, $10$ in Ant-Dir, Cheetah-vel, and Cheetah-Dir tasks with reward change, $20$ in Hopper-Param and Walker-Param tasks with dynamics change.

\item The context encoder is modeled as multi-layer perceptrons(MLPs) with 3 hidden layers with the final liner layer to produce a context embedding in $64$ dimensions. The generalized reward model $R_\psi$ and dynamics model $P_\omega$ both shared with the structure of the first half of the U-Net used for the diffusion model, with a final linear layer to produce a scalar output. 

\item We jointly train the context encoder and corresponding reward model and dynamics model using the Adam optimizer~\citep{ba2015adam} with a learning rate of \num{2e-4} and batch size of $64$ for 1000 epochs.

\item Timestep and context embeddings, both $64$-dimensional vectors, are produced by separate 2-layered MLP (with $256$ hidden units and Mish nonlinearity) and are concatenated together before getting added to the activations of the first temporal convolution within each block. We borrow the code for temporal U-Net from the official implementation of Diffuser\footnote{\scriptsize\url{https://github.com/jannerm/diffuser}}. 


\item We train noise model $\epsilon_\theta$ using the Adam optimizer~\citep{ba2015adam} with a learning rate of \num{2e-4} and batch size of $32$ for \num{1e6} train steps.


\item We choose the proper probability $\beta$ of removing the conditioning information for each task, the detailed choice, and the results can be found in ~\cref{abla:drop}.

\item We use $k \in \{20,50,100\}$ diffusion steps.

\item We use a planning horizon $H$ of $4$ in Point-Robot task, $16$ in Cheetah-Vel and Cheetah-Dir tasks, $32$ in Ant-Dir, Hopper-Param and Walker-Param tasks.

\item We use a guidance scale $\omega \in \{1.2, 1.4, 1.6, 1.8, 2.0\}$ but the exact choice varies by task.

\end{itemize}

\section{Details about Pseudocode for Implementation}
\label{app:torch-like}
Here, we provide pseudocode for our implementation.

\begin{algorithm}[htb]
\caption{\alg Training Pytorch-like Pseudocode}
\definecolor{codeblue}{rgb}{0.28125,0.46875,0.8125}
\lstset{
    basicstyle=\fontsize{9pt}{9pt}\ttfamily\bfseries,
    commentstyle=\fontsize{9pt}{9pt}\color{codeblue},
    keywordstyle=
}
\begin{lstlisting}[language=python]

def training(trajectory, context_drop_prob):
  """
  trajectory: expert trajectory from offline data in training set
  context_drop_prob: the probability to drop conditional context
  """
  
  # Capture task information
  sub_trajs = random_sample(trajectory, len=h)  
  # extract sub-segement from whole trajectory to infer context, h denotes historical length
  latent_embeds = []
  for sub_traj in sub_trajs:
    latent_embed = context_encoder(sub_traj)
    latent_embeds.append(latent_embed)
  context_embed = aggreate(latent_embeds) # arbitrary aggreation technique 

  mask_dist = Bernoulli(probs=context_drop_prob) 
  mask = mask_dist.sample()
  masked_context_embed = context_embed * mask
  
  # sample sub-traj for diffuison model to estimate
  traj = random_sample(trajectory, len=H) # H denotes planning horizon

  # Diffuse process on traj
  t = randint(0, T) # time step
  eps = normal(mean=0, std=1)  
  nosiy_traj = sqrt(alpha_cumprod(t)) * traj + sqrt(1 - alpha_cumprod(t)) * eps

  # Denoise process to reconstruct traj
  pred_traj = denoise_model(noisy_traj, masked_context_embed, t)

  # Set prediction loss
  loss = set_prediction_loss(pred_traj, noisy_traj)
  return loss

\end{lstlisting}
\end{algorithm}

\begin{algorithm}[t]
\caption{\alg Sampling Pytorch-like Pseudocode}
\definecolor{codeblue}{rgb}{0.28125,0.46875,0.8125}
\lstset{
    basicstyle=\fontsize{9pt}{9pt}\ttfamily\bfseries,
    commentstyle=\fontsize{9pt}{9pt}\color{codeblue},
    keywordstyle=
}

\begin{lstlisting}[language=python]

def Sampling(warm_start_data, steps, T):
  """
  warm_start_data: the provided trajectories collected in the test task
  # steps: number of sample steps
  # T: number of time steps
  """

  # Capture task information
  sub_trajs = random_sample(warm_start_data, len=h)  
  # extract sub-segement from warm_start_data to infer context, h denotes historical length
  latent_embeds = []
  for sub_traj in sub_trajs:
    latent_embed = context_encoder(sub_traj)
    latent_embeds.append(latent_embed)
  context_embed = aggreate(latent_embeds)
 
  # noisy traj to be denoised, length set as planning horizon H 
  traj_t = normal(mean=0, std=1)

  # uniform sample step size
  times = reversed(linespace(-1, T, steps))
  
  # [(T-1, T-2), (T-2, T-3), ..., (1, 0), (0, -1)]
  time_pairs = list(zip(times[:-1], times[1:])

  for t_now, t_next in zip(time_pairs):
    # Predict traj from pb_t
    pred_traj = denoise_model(traj_t, context_embd, t_now)
    
    # Estimate pb_t at t_next
    traj_t = denoise_step(traj_t, pred_traj, t_now, t_next)
    
  return pred_traj

\end{lstlisting}
\end{algorithm}


\end{document}